\documentclass{article}

\usepackage[table]{xcolor}
\usepackage{float}
\usepackage[preprint]{corl_2026} 
\usepackage{booktabs}
\usepackage{graphicx}
\usepackage{wrapfig}
\usepackage{amsmath}
\usepackage{amssymb}
\usepackage{array}
\usepackage{tabularx}
\usepackage{placeins}
\usepackage{multirow}

\floatstyle{ruled}
\newfloat{algorithm}{t}{loa}
\floatname{algorithm}{Algorithm}


\definecolor{spacevlnBlueRow}{HTML}{EAF4FF}
\definecolor{spacevlnGreenRow}{HTML}{EAF8EF}
\definecolor{spacevlnPurpleRow}{HTML}{F3ECFF}
\definecolor{spacevlnGrayRow}{HTML}{F4F6F8}
\definecolor{spacevlnOwnRow}{HTML}{F1F3F5}
\definecolor{spacevlnBlueText}{HTML}{1F5FAD}
\definecolor{spacevlnGreenText}{HTML}{2F7D55}
\definecolor{spacevlnPurpleText}{HTML}{7A3EC8}

\newcommand{\cmark}{\ensuremath{\checkmark}}
\newcommand{\xmark}{\ensuremath{\times}}

\newcommand{\best}[1]{\textbf{#1}}
\newcommand{\second}[1]{\underline{#1}}
\newcolumntype{L}[1]{>{\raggedright\arraybackslash}p{#1}}
\newcolumntype{C}[1]{>{\centering\arraybackslash}p{#1}}

\title{SpaceVLN: A Zero-Shot Vision-and-Language Navigation Agent with Online Spatial Cognitive Memory and Reasoning}

\hypersetup{
  pdftitle={SpaceVLN: A Zero-Shot Vision-and-Language Navigation Agent with Online Spatial Cognitive Memory and Reasoning},
  pdfauthor={Yucheng Deng, Pingrui Lai, Xinhai Li, Chenjia Bai, Xiaoheng Deng, Chengnuo Sun, Xuelong Li, Hua Yang},
  pdfsubject={Preprint}
}

\author{
  \textbf{Yucheng Deng$^{1,2}$, Pingrui Lai$^1$, Xinhai Li$^{2,\ddagger}$, Chenjia Bai$^2$,}\\
  \textbf{Xiaoheng Deng$^3$, Chengnuo Sun$^{2,4}$, Xuelong Li$^{2,\dagger}$, Hua Yang$^{1,\ddagger\dagger}$}\\[0.35em]
  {\normalfont $^1$School of Information Science and Electronic Engineering \& School of Integrated Circuits,}\\
  {\normalfont Shanghai Jiao Tong University}\\
  {\normalfont $^2$Institute of Artificial Intelligence, China Telecom}\\
  {\normalfont $^3$Central South University \qquad $^4$Jiangsu University}
}

\begin{document}
\maketitle
\begingroup
\renewcommand{\thefootnote}{\fnsymbol{footnote}}
\footnotetext[3]{Project leaders. \quad $^\dagger$ Corresponding authors.}
\endgroup


\begin{abstract}
Vision-and-Language Navigation in continuous environments requires agents to
understand the spatial structure of unseen environments in order to follow
language instructions.
Although foundation models have opened a promising path toward zero-shot navigation without task-specific policy training, many navigators still rely on local visual cues and linear history-based reasoning, overlooking the spatial nature of navigation across explored regions, traversed paths, landmarks, and their spatial relations.
In this paper, we propose SpaceVLN, a navigation agent built around Spatial Cognitive Memory and Task-Guided Spatial Reasoning.
Specifically, SpaceVLN introduces an efficient stagewise closed-loop framework where planning and execution are organized around verifiable space--landmark stages. 
During navigation, the agent progressively abstracts explored regions into Spatial Waypoints and dynamically maintains subtask-grounded landmark evidence, forming a hierarchical Spatial Cognitive Memory for progress localization and spatial-relation understanding. 
Built on this memory, Spatial-CoT integrates task-progress reasoning with spatial perception, analysis, and prediction, enabling Task-Guided Spatial Reasoning for embodied navigation.
The unified stage interface enables SpaceVLN to address both Vision-and-Language Navigation and Object-Goal Navigation under a unified zero-shot setting, without task-specific policy training.
Across R2R-CE, RxR-CE, GN-Bench, and HM3D-OVON, SpaceVLN achieves state-of-the-art zero-shot performance, and real-robot deployment further validates its applicability.
These results highlight Spatial Cognitive Memory and Task-Guided Spatial Reasoning as a practical foundation for stronger embodied navigation agents. \href{https://charles-donne.github.io/SpaceVLN/}{\textbf{Project page}.}
\end{abstract}



\section{Introduction}

Vision-and-Language Navigation (VLN) requires an embodied agent to follow
natural-language instructions in unseen 3D environments.
Early benchmarks such as R2R and RxR define navigation on discrete graphs, whereas VLN in continuous
environments (VLN-CE) requires primitive motion in metric
space~\citep{Anderson2018VLN,Krantz2020VLNCE,Ku2020RxR}. This setting is
closer to physical deployment and demands stronger spatial ability: a 
navigation agent cannot decide actions from local observations alone, but must
understand spatial structure to follow instructions and complete navigation.

Pre-built map-based navigators provide spatial context but typically require
costly pre-exploration scene representations~
\citep{Bhatt2025VLNZero,Zhang2026SpatialNav,Zhang2026SpatialAnt}.
Foundation models enable zero-shot navigation without task-specific policy
training, while broader AI-flow perspectives emphasize resource-aware
coordination of large models across device, edge, and cloud systems~\citep{An2025AIFlow}.
Recent foundation-model navigators strengthen VLM reasoning with progress
estimation, graph constraints, history analysis, and future prediction~
\citep{Zhou2024NavGPT,Qiao2025OpenNav,Chen2025CANav,Yin2025GCVLN,Dai2026EvoNav}.
Yet their reasoning remains largely tied to local observations and linear
histories, limiting spatial cognition and making agents easily lost in
multi-space or complex navigation tasks.
\begin{figure}[t!]
    \centering
    \includegraphics[width=\linewidth]{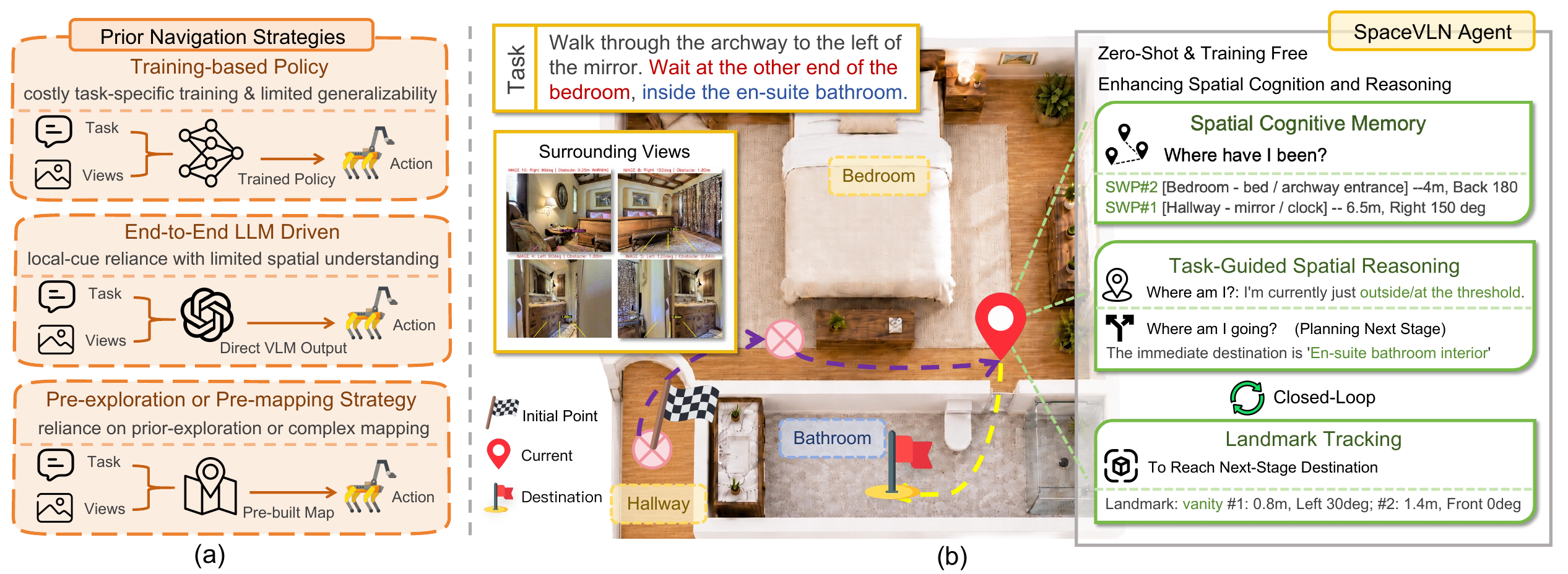}
    \caption{\textbf{Comparison between prior navigation strategies and
    SpaceVLN.}
    \textbf{(a)} Prior strategies rely on task-specific training, direct
    VLM action prediction, or pre-built maps, limiting generalization and
    weakening spatial progress grounding.
    \textbf{(b)} SpaceVLN is a zero-shot, training-free navigation agent.
    It leverages online Spatial Cognitive Memory and
    Task-Guided Spatial Reasoning to strengthen the agent's spatial cognition
    and reasoning ability, thereby improving navigation success.}
    \label{fig:teaser}
\end{figure}

We introduce \textbf{SpaceVLN}, a zero-shot navigation agent that strengthens
spatial perception, memory, and reasoning in unseen environments.
SpaceVLN constructs online Spatial Cognitive Memory by abstracting surrounding observations and landmark detections within explored regions.
A planner uses this memory to localize task progress and generate the next
verifiable space--landmark stage, while an efficient executor executes the stage 
using local visual cues, landmark observations, and obstacle evidence. 
By converting different tasks into verifiable space--landmark stages, the same
framework supports diverse navigation tasks without task-specific retraining or
adaptation. After each stage, the agent checks progress from new observations
and memory updates, then revises or replans the next stage rather than following
a fixed open-loop plan.

In summary, our main contributions are: 
    (i) We propose \textbf{SpaceVLN} with an efficient stagewise closed-loop navigation framework that
    organizes planning and execution around verifiable space--landmark stages,
    enabling unified task handling, efficient execution, and self-correction 
    across Vision-and-Language Navigation and Object-Goal Navigation.
    (ii) We introduce \textbf{Spatial Cognitive Memory}, which progressively
    abstracts explored regions into Spatial Waypoints and dynamically maintains
    subtask-grounded landmark cues, strengthening spatial cognition and spatial-relation understanding.
    (iii) We propose \textbf{Spatial-CoT}, a schema-guided chain-of-thought that couples task-chain progress with spatial perception, spatial analysis, and spatial prediction,
    forming a Task-Guided Spatial Reasoning mechanism for embodied navigation. 
Extensive evaluations across Vision-and-Language Navigation, Object-Goal
Navigation, component ablations, and real-robot deployment demonstrate the
effectiveness of SpaceVLN. SpaceVLN achieves state-of-the-art zero-shot SR of
53.3, 48.9, 39.3, and 51.6 on R2R-CE, RxR-CE, GN-Bench, and HM3D-OVON.


\section{Related Work}
\label{sec:related work}

\subsection{Vision-and-Language Navigation}
\label{sec:related-vln}

Vision-and-Language Navigation (VLN) requires an embodied agent to follow
natural-language route instructions in unseen 3D environments. R2R and RxR
formulate VLN on discrete viewpoint graphs, whereas VLN in continuous
environments (VLN-CE) removes the predefined graph and requires low-level
control in continuous metric
space~\citep{Anderson2018VLN,Ku2020RxR,Krantz2020VLNCE}. This setting better
approximates physical deployment and requires agents to ground instruction
progress in spatial structure. Training-based methods, mainly based on
imitation or reinforcement learning, have incorporated pre-training, physical
priors, waypoint prediction, map/graph reasoning, memory-observation fusion,
and future
prediction~\citep{Fried2018SpeakerFollower,Hao2020PREVALENT,Chen2021HAMT,Chen2022DUET,Hong2022DiscreteContinuousVLN,An2023BEVBert,An2025ETPNav,Yu2025MossVLN,Zhang2025MapNav,Zhang2024NaVid,Zhang2025UniNaVid}.
However, they usually require large amounts of task-specific navigation data,
learned waypoint modules, or supervised policies, and many focus on local
action or waypoint prediction without maintaining a global spatial state.

Pre-built scene representations provide spatial context, but usually
require costly pre-exploration or static offline
memory~\citep{Bhatt2025VLNZero,Zhang2026SpatialNav,Zhang2026SpatialAnt,Zhou2025FSRVLN}.
Foundation-model agents enable zero-shot navigation without training 
and improve reasoning with progress estimation, graph
constraints, history analysis, and future prediction~\citep{Zhou2024NavGPT,Chen2024MapGPT,Qiao2025OpenNav,Chen2025CANav,Yin2025GCVLN,Dai2026EvoNav,Li2026GTA}.
However, many still rely on linear memory and reasoning, with limited
perception and analysis of 3D spatial structure, making agents prone to getting
lost.
Although DeCoNav introduces online region-level semantic states for coordination, it
relies on trained modules and lacks a
clear hierarchical spatial structure~\citep{Zhou2026DeCoNav}. This paper
focuses on training-free spatial understanding in complex continuous
environments.

\subsection{Foundation Models for Embodied Navigation and Spatial Reasoning}
\label{sec:related-foundation-navigation}

Foundation models offer strong high-level reasoning, but they do not
inherently account for physical affordances, motion feasibility, or real-time
environmental feedback. Recent robotics systems therefore adopt hybrid
architectures that use language-capable models as translators, checkers, or
planners, while connecting their outputs to executable behavior through value
maps, task-and-motion constraints, or 3D scene
graphs~\citep{Huang2022ZeroShotPlanners,Ahn2022SayCan,Yao2023ReAct,Lin2023Text2Motion,Huang2023VoxPoser,Rana2023SayPlan,Chen2024AutoTAMP,Rajvanshi2024SayNav}.
Navigation agents follow the same pattern by providing foundation models with
structured environment context, such as maps, progress estimates, graph
constraints, or metric world states, or by using the model for high-level
planning while leaving low-level control to an execution
module~\citep{Chen2024MapGPT,Qiao2025OpenNav,Chen2025CANav,Yin2025GCVLN,Li2026GTA}.

At the same time, spatial understanding remains a central bottleneck for
embodied foundation models. OpenEQA reports gaps on questions about physical
environments~\citep{Majumdar2024OpenEQA}, Thinking in Space finds local spatial
awareness but weak global spatial reasoning in VLMs~\citep{Yang2025ThinkingInSpace},
and NavSpace distinguishes spatially grounded Vision-and-Language Navigation
from generic semantic navigation~\citep{Yang2025NavSpace}. SpaceVLN follows
this hybrid-agent view~\citep{Pan2026RobotNavigationSurvey} by treating
foundation models as replaceable reasoning backends, providing them with richer
online spatial context, and casting their outputs as typed navigation stages
that can be executed, verified, and revised through closed-loop feedback.
	

\section{Method}
\label{method}
\label{sec:method}

\begin{figure}[t]
    \centering
    \includegraphics[width=0.90\linewidth]{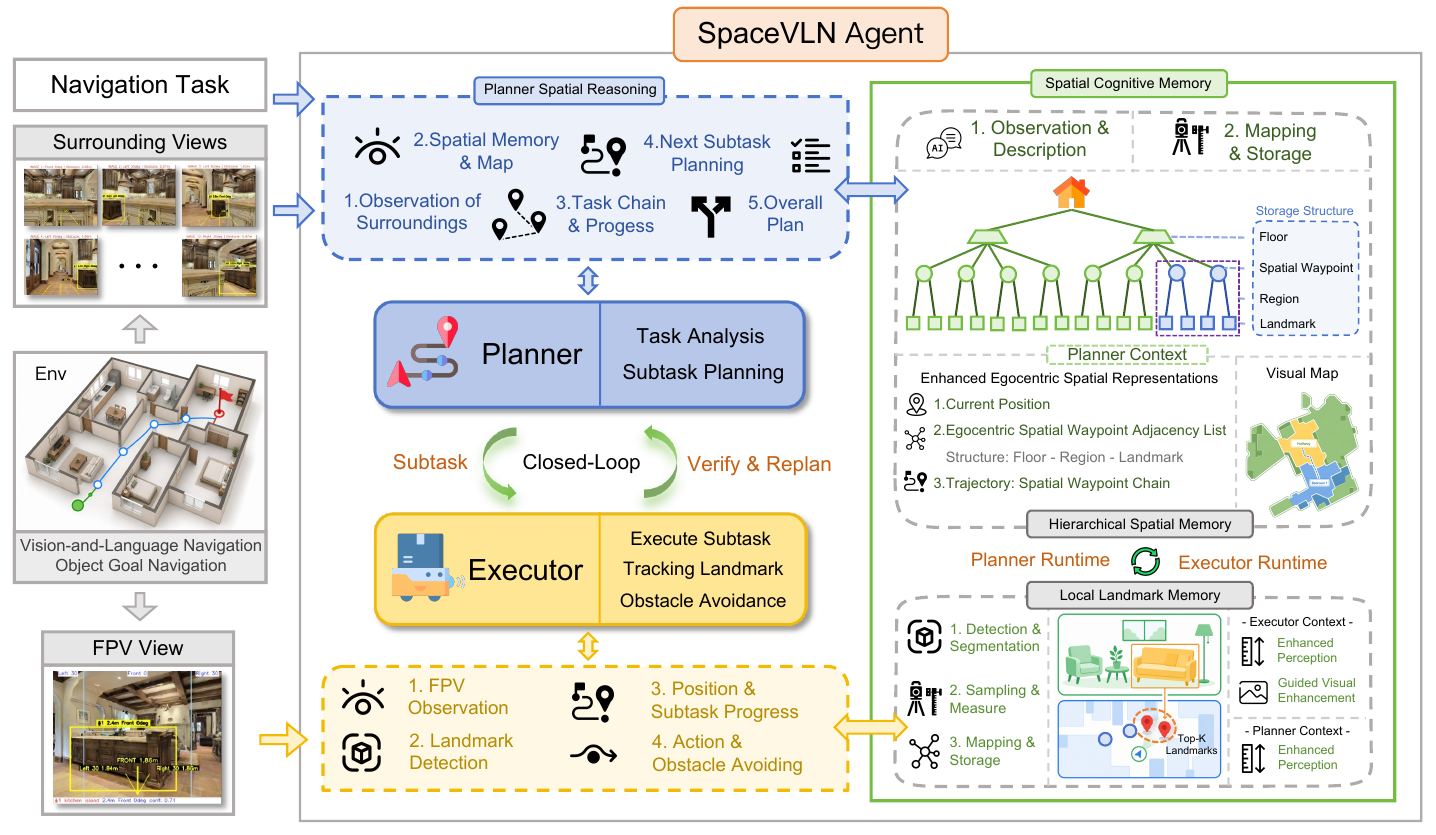}
   \caption{\textbf{Framework of SpaceVLN.}
SpaceVLN realizes navigation through a stagewise closed-loop framework, augmented by Spatial Cognitive Memory and Task-Guided Spatial Reasoning.
The planner infers progress from 12-direction look-around and spatial memory, then generates a verifiable space--landmark stage as an executable subtask.
The executor follows this subtask using FPV views and landmark memory, while closed-loop updates memory for verification and replanning.}
\label{fig:framework}
\end{figure}

Zero-shot embodied navigation in unseen environments requires
an agent to receive a navigation task $q$, online RGB-D observations, and pose
estimates, and issue primitive actions without task-specific policy training.
We develop \textbf{SpaceVLN} as a stagewise
closed-loop navigation framework built on Spatial Cognitive Memory and
Task-Guided Spatial Reasoning. Figure~\ref{fig:framework} gives the overall
framework; Secs.~\ref{sec:method-stagewise}--\ref{sec:method-reasoning}
describe its main components, while Appendix~\ref{app:method-overview} and
Fig.~\ref{fig:app-reasoning-pipeline} provide the runtime pipeline and VLM
context architecture.

\subsection{Stagewise Closed-Loop Navigation Framework}
\label{sec:method-stagewise}

This section introduces the stagewise closed-loop navigation framework of
SpaceVLN. The framework coordinates planning and execution through verifiable
space--landmark stages: at planning step $k$, the planner reads the task,
current observations, and Spatial Cognitive Memory $\mathcal{M}_k$ to produce
an executable subtask; the executor then acts over local time steps $t$ until
control returns for verification or replanning. We define the online memory
state in Sec.~\ref{sec:method-memory}.

\paragraph{Planner.}
For navigation stage $k$, the planner receives the task $q$, Spatial Cognitive Memory $\mathcal{M}_k$, and a 12-view look-around
$\mathcal{O}^{12}_{k}=\{(o^i_{k},\alpha_i)\}_{i=1}^{12}$, where
$o^i_{k}$ is the image view along heading $\alpha_i$. It constructs a
task-conditioned space--landmark anchor chain:
\begin{equation}
    \mathcal{C}_k=(v^0_k,\ldots,v^{N_k}_k),\qquad
    v^i_k=(s^i_k,\lambda^i_k),
\end{equation}
where each anchor pairs a spatial state $s^i_k$ (e.g., a room, connector, or
floor transition) with an identifying cue $\lambda^i_k$ (e.g., a landmark,
object, or entrance). The planner matches the current state to an anchor
$v^{j_k}_k$ in this chain and takes the immediate next unverified anchor
$v^{j_k+1}_k$ as the stage target when such an anchor remains.

\paragraph{Stage-Level Subtask.}
Rather than providing a full route, planner outputs only the next
stage-level subtask:
\begin{equation}
    \sigma_k=(v^{j_k}_k,\mathcal{C}_k,v^{j_k+1}_k,\delta_k,u_k,\ell_k,z_k),
    \qquad \delta_k\in\Delta^{12}_{k},\quad j_k<N_k,
    \label{eq:stage-subtask}
\end{equation}
where $v^{j_k}_k$ and $v^{j_k+1}_k$ are the current and next anchors,
$\mathcal{C}_k$ is the anchor chain, $\delta_k$ is the selected direction,
$u_k$ is the stage instruction passed to the executor, $\ell_k$ is the subtask
landmark, and $z_k$ is the completion flag. If $j_k=N_k$, the planner sets
$z_k=1$. Thus, $\sigma_k$ binds anchor, instruction, and landmark into a task-adaptive,
executable, and verifiable stage representation.

\paragraph{Executor.}
The executor follows the current subtask using FPV views and environment
perception:
\begin{equation}
    a_t=\pi_{\mathrm{E}}(I_t,\Omega_t,\Lambda_t,\sigma_k),
    \qquad t_k\leq t<t_{k+1}.
\end{equation}
where $\Omega_t$ and $\Lambda_t$ denote obstacle and subtask landmark
perception. Rather than directly mapping images to actions, the executor
performs local control toward the next anchor using the subtask landmark and
traversability cues. Control returns to the planner upon destination arrival,
task completion, stage-budget exhaustion, or blockage:
\begin{equation}
    \mathcal{F}_k=\Phi(\sigma_k,\tau_k,\mathcal{B}^{(k)}),\qquad
    \sigma_{k+1}=\pi_{\mathrm{P}}
    (q,\mathcal{O}^{12}_{k+1},\mathcal{M}_{k+1},\mathcal{F}_k).
\end{equation}
Detailed pipeline, interfaces, and safeguards are provided in
Appendices~\ref{app:method-overview}, \ref{app:stagewise-policy},
and~\ref{app:reliability-details}.

\subsection{Spatial Cognitive Memory}
\label{sec:method-memory}

For embodied navigation, memory should organize online perception into a
relational spatial state rather than a flat observation cache. Such a state
summarizes explored regions, traversed paths, landmarks, and their spatial
connectivity. Hierarchical scene graphs and spatial maps have made large
environments queryable for language
reasoning~\citep{Rana2023SayPlan,Werby2024HOVSG,Zhang2026SpatialNav,Zhou2025FSRVLN},
but typically assume pre-explored or pre-built environment representations.
SpaceVLN instead builds \emph{Spatial Cognitive Memory} online as a shared
planner--executor state:
\begin{equation}
    \mathcal{M}_t=\{\mathcal{W}_t,\mathcal{L}_t\}, \qquad
    \mathcal{W}_t=\{\mathcal{G}_t,\mathcal{P}_t\},
\end{equation}
where $\mathcal{W}_t$ denotes Hierarchical Spatial Memory, including the
Spatial Waypoint graph $\mathcal{G}_t$ and executed Spatial Waypoint chain
$\mathcal{P}_t$, while $\mathcal{L}_t$ denotes Local Landmark Memory.
This representation exposes spatial context to the VLM while preserving
relations among floors, regions, Spatial Waypoints, and landmarks.
\begin{figure}[t]
    \centering
    \includegraphics[width=\linewidth,trim=0pt 4pt 0pt 4pt,clip]{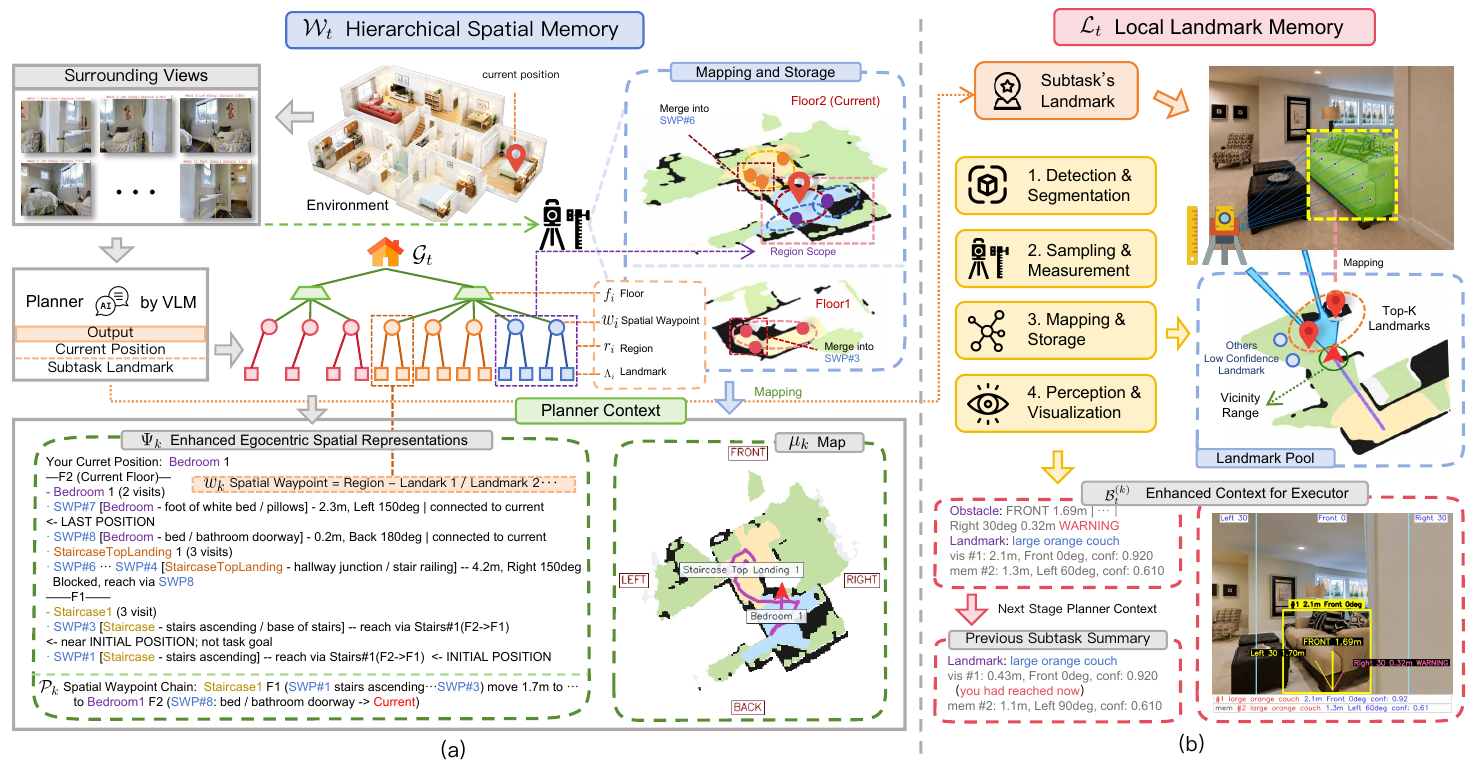}
    \caption{\textbf{Spatial Cognitive Memory.}
    (a) Hierarchical Spatial Memory abstracts views and traversed paths into a
    Spatial Waypoint graph and executed Spatial Waypoint chain, then converts
    them into enhanced egocentric spatial representation for planner.
    (b) Local Landmark Memory maintains a subtask-specific landmark pool,
    providing top-$K$ cues for execution and next-stage planning.}
    \label{fig:memory-detail}
\end{figure}

\paragraph{Hierarchical Spatial Memory.}
The persistent unit is the \emph{Spatial Waypoint}. SpaceVLN organizes these
waypoints into a graph and an executed Spatial Waypoint chain:
\begin{equation}
    \mathcal{G}_t=(\{w_i\}_{i=1}^{N_t},\mathcal{E}_t), \qquad
    w_i=(x_i,f_i,r_i,\Lambda_i).
\end{equation}
where $x_i$ is the metric pose, $f_i$ is the floor, $r_i$ is the region, and
$\Lambda_i$ denotes local landmark cues. The edge set $\mathcal{E}_t$ captures
region membership and traversability, preserving spatial
topology without dense reconstruction or pre-exploration.

At each planning step $k$, SpaceVLN serializes the hierarchy around the agent's
current position:
\begin{equation}
    \Psi_k = \operatorname{Serialize}
    \left(\operatorname{Arrange}_{\mathcal{G}_k}(w_k;w_0,w_{k-1}),
    \mathcal{P}_k,\mu_k\right),
\end{equation}
where $w_k$, $w_0$, and $w_{k-1}$ denote the current, initial, and previous
Spatial Waypoints. $\operatorname{Arrange}_{\mathcal{G}_k}$ merges overlapping
waypoints and orders all Spatial Waypoints around $w_k$ into an agent-centered
adjacency list with known connectivity paths; $\mathcal{P}_k$ and $\mu_k$
provide the executed Spatial Waypoint chain and rendered map. The resulting
$\Psi_k$ is the enhanced egocentric spatial representation used by the planner.
Appendix~\ref{app:memory-details} gives the graph construction and
serialization details.

\paragraph{Local Landmark Memory.}
Hierarchical Spatial Memory provides global spatial context, while execution
requires local landmark details for action generation. 
Given the subtask landmark $\ell_k$, SpaceVLN maintains a ranked landmark pool through online grounding:
\begin{equation}
    \mathcal{B}^{(k)}_t=
    \operatorname{TopK}\{(c_i,\bar{x}_i,d_i,\beta_i,\gamma_i)
    : c_i\in\operatorname{Norm}(\ell_k)\}.
\end{equation}
Each entry stores category $c_i$, fused position $\bar{x}_i$, distance $d_i$,
bearing $\beta_i$, and confidence $\gamma_i$. The executor combines FPV
observations, the top-$K$ landmark candidates, and obstacle cues to execute
primitive actions, and the updated landmark state is returned to the planner
for the next stage.

\subsection{Task-Guided Spatial Reasoning}
\label{sec:method-reasoning}

\begin{figure}[t!]
    \centering
    \includegraphics[width=\linewidth]{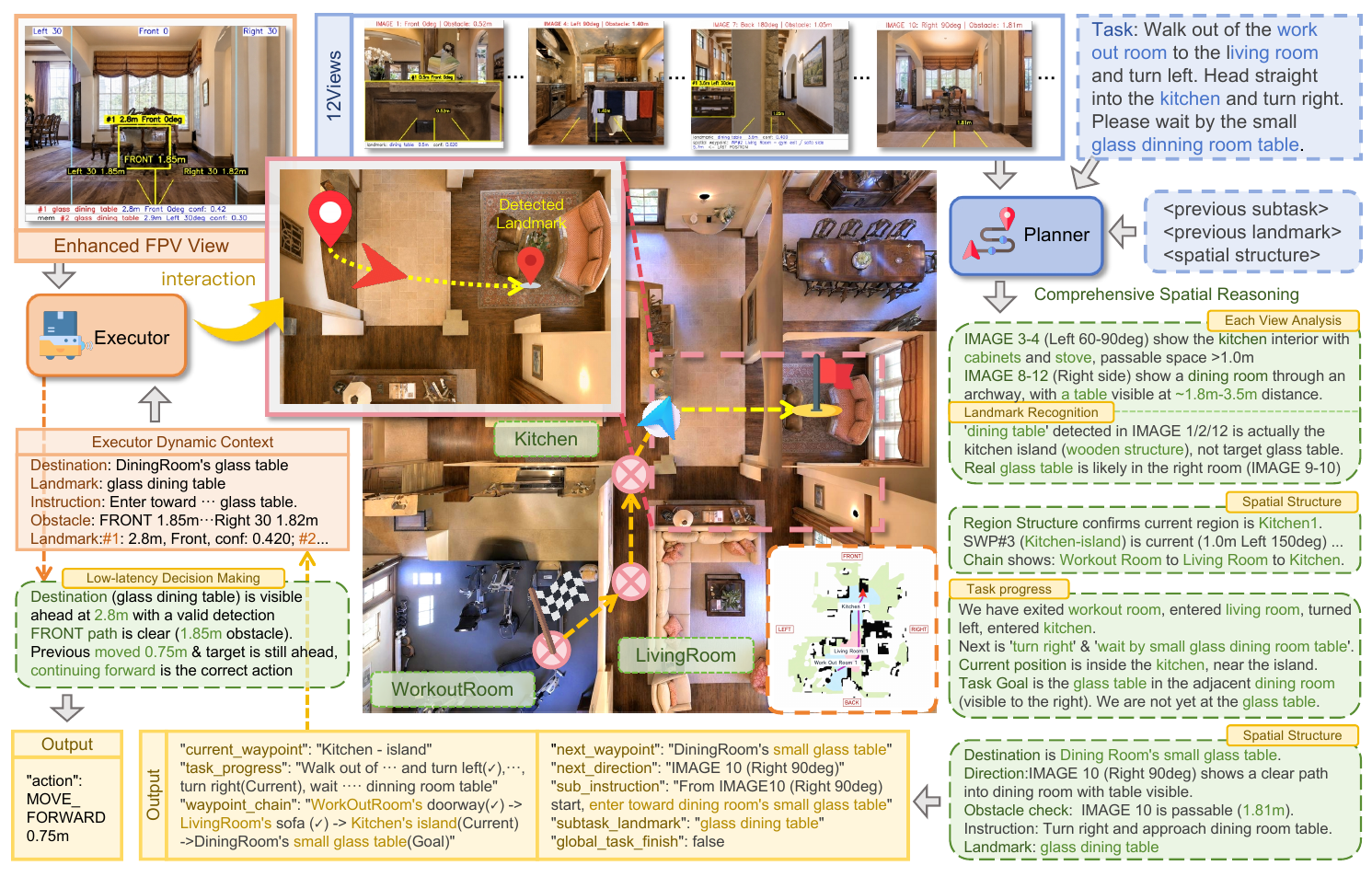}
    \caption{\textbf{Task-Guided Spatial Reasoning Example.}
    Given an instruction and a 12-view look-around, the planner follows
    Spatial-CoT for environment analysis, localization, and planning, then
    outputs a structured stage-level subtask. The executor combines the enhanced
    FPV view, landmark detection, and obstacle cues to complete the subtask and
    return feedback for next-stage planning.}
    \label{fig:reasoning-example}
\end{figure}

Recent foundation-model navigators expose intermediate reasoning to improve
direct action prediction, including explicit reasoning traces in
NavGPT~\citep{Zhou2024NavGPT}, spatial-temporal progress estimation in
Open-Nav~\citep{Qiao2025OpenNav}, and history-to-future reasoning in
EvoNav~\citep{Dai2026EvoNav}. 
SpaceVLN extends this direction by incorporating spatial memory into planner reasoning.
We introduce \emph{Spatial-CoT}, a schema-guided spatial variant of CoT~\citep{Wei2022CoT,Kojima2022ZeroShotCoT,Li2023StructuredCoT},
where an instruction-tuned VLM outputs structured fields conditioned on task
progress and spatial memory.
Figure~\ref{fig:reasoning-example} illustrates this
reasoning process in a concrete episode, and
Appendix~\ref{app:output-contracts} provides the reasoning details and
Spatial-CoT schemas.

\paragraph{Planner reasoning.}
Let $\Psi_k$ denote the enhanced egocentric spatial representation from
Section~\ref{sec:method-memory}. 
At planning step $k$, the planner performs Spatial-CoT reasoning:
\begin{equation}
    r_k^{\mathrm{P}}=
    [r_{\mathrm{obs}}, r_{\mathrm{mem}}, r_{\mathrm{prog}},
    r_{\mathrm{dec}}, r_{\mathrm{plan}}],
\end{equation}
covering observation analysis, perception grounded in Spatial Cognitive Memory,
progress estimation, next-stage planning, and plan summarization. 
The localization and progress fields verify the task anchor chain against the
12-view observation and spatial context:
\begin{equation}
    j_k =
    \max\left\{i:\operatorname{Verify}
    (v^i_k\mid \mathcal{O}^{12}_{k},\Psi_{k},\mathcal{P}_{k})=1
    \right\},
\end{equation}
This identifies the furthest verified anchor $v^{j_k}_k$ and parses
$r_k^{\mathrm{P}}$ into the next subtask $\sigma_k$ in
Eq.~\ref{eq:stage-subtask}.

\paragraph{Executor reasoning.}
Given the current stage-level subtask $\sigma_k$, the executor restricts
reasoning to local action generation rather than route revision. 
It checks the active landmark and stage progress, then generates an admissible
primitive action through the executor-level Spatial-CoT schema:
\begin{equation}
    r_t^{\mathrm{E}}=
    [r_{\mathrm{view}},r_{\mathrm{det}},r_{\mathrm{prog}},r_{\mathrm{act}}],
    \qquad
    a_t=\operatorname{Parse}_{\mathcal{U}_t}(r_t^{\mathrm{E}}),
\end{equation}
where the four fields cover front-view content, landmark validation,
stage progress, and obstacle-aware action generation, and
$\mathcal{U}_t\subseteq\mathcal{A}$ is the locally admissible action set. 
This separates stage-level spatial reasoning from high-frequency executor control. 
\FloatBarrier

\section{Experiments}
\label{sec:result}

\subsection{Experiment Setup}
\label{sec:experiment-setup}

\paragraph{Benchmarks.}
We evaluate SpaceVLN on three Vision-and-Language Navigation benchmarks:
R2R-CE and RxR-CE Val-Unseen~\citep{Anderson2018VLN,Ku2020RxR,Krantz2020VLNCE},
and GN-Bench, which features more complex instructions. We further evaluate HM3D-OVON
Val-Unseen for open-vocabulary Object-Goal Navigation~\citep{Yokoyama2024HM3DOVON}.
R2R-CE and RxR-CE are implemented in the Habitat simulator over Matterport3D
scenes~\citep{Savva2019Habitat,Chang2017Matterport3D,Krantz2020VLNCE}; we
evaluate all 1,839 R2R-CE Val-Unseen episodes and the 2,433-episode en-US
subset of RxR-CE Val-Unseen. HM3D-OVON uses HM3D
scenes, and we evaluate its 3,000-episode Val-Unseen split~\citep{Ramakrishnan2021HM3D,Yokoyama2024HM3DOVON}.
Appendix~\ref{app:experiment-setup} gives detailed splits and reporting rules.

\paragraph{Metrics.}
Following prior VLN-CE evaluations, we use success rate (SR) and success
weighted by path length (SPL) as the primary metrics. SR measures whether
the agent stops within 3m of the target, while SPL additionally accounts for
path efficiency. We also report navigation error (NE) and oracle success rate
(OSR) for VLN-style benchmarks, and SR/SPL for HM3D-OVON.

\paragraph{Deployment.}
All simulator experiments share a unified architecture and controller interface,
with egocentric RGB-D observations and metric poses as inputs and primitive
actions as outputs.
Main results use Qwen3.5-Plus for planning and Qwen3.5-Flash for execution.
Spatial mapping is built from RGB-D
observations, and landmark grounding is implemented with GroundingDINO and
RepViT-SAM~\citep{Liu2023GroundingDINO,Wang2023RepViTSAM}.
We further test SpaceVLN on a TX-Q1 differential-drive robot with a RealSense
D435i RGB-D camera in 50 custom indoor navigation episodes.
Real-robot deployment results and analysis, together with model and runtime
comparisons, are reported in Appendices~\ref{app:real-robot-details}
and~\ref{app:runtime-cost}.

\subsection{Main Results}
\label{sec:experiment-main-results}

\paragraph{Vision-and-Language Navigation.}
\begin{table}[t!]
\centering
\scriptsize
\setlength{\tabcolsep}{2.0pt}
\renewcommand{\arraystretch}{1.04}
\caption{Comparison on R2R-CE and RxR-CE Val-Unseen in continuous environments.
\textbf{Best} and \underline{second} results are highlighted.}
\label{tab:r2rce_rxrce_merged}
\begin{tabular}{@{}L{0.27\textwidth}|*{4}{C{0.070\textwidth}}|*{4}{C{0.070\textwidth}}@{}}
\toprule
\multirow{2}{=}{\textbf{Method}} &
\multicolumn{4}{c|}{\begin{tabular}{@{}c@{}}\textbf{R2R-CE Val-Unseen}\\[-3.1pt]\rule{0.26\textwidth}{0.3pt}\end{tabular}} &
\multicolumn{4}{c@{}}{\begin{tabular}{@{}c@{}}\textbf{RxR-CE Val-Unseen}\\[-3.1pt]\rule{0.26\textwidth}{0.3pt}\end{tabular}} \\
& \textbf{NE}$\downarrow$ & \textbf{OSR}$\uparrow$ & \textbf{SR}$\uparrow$ & \textbf{SPL}$\uparrow$
& \textbf{NE}$\downarrow$ & \textbf{OSR}$\uparrow$ & \textbf{SR}$\uparrow$ & \textbf{SPL}$\uparrow$ \\
\midrule
\rowcolor{spacevlnBlueRow}
\multicolumn{9}{@{}l}{\textbf{\textit{Supervised learning}}} \\
CMA~\citep{Krantz2020VLNCE} & 6.3 & 49.0 & 38.0 & 33.0 & 10.4 & -- & 24.1 & 19.1 \\
BEVBert~\citep{An2023BEVBert} & 4.7 & 67.0 & 59.0 & 50.0 & 4.8 & -- & 64.4 & -- \\
NaVid~\citep{Zhang2024NaVid} & 5.5 & 49.1 & 37.4 & 35.9 & 8.4 & 34.5 & 23.8 & 21.2 \\
ETPNav~\citep{An2025ETPNav} & 4.7 & 65.0 & 57.0 & 49.0 & 5.6 & -- & 54.8 & 44.9 \\
Uni-NaVid~\citep{Zhang2025UniNaVid} & 5.6 & 53.3 & 47.0 & 42.7 & 6.2 & 55.5 & 48.7 & 40.9 \\
\midrule
\rowcolor{spacevlnBlueRow}
\multicolumn{9}{@{}l}{\textbf{\textit{Pre-exploration}}} \\
VLN-Zero~\citep{Bhatt2025VLNZero} & 6.0 & 51.6 & 42.4 & 26.3 & 9.1 & -- & 30.8 & 19.0 \\
EvoNav (Gemini-2.5-Pro)~\citep{Dai2026EvoNav} & 5.0 & 51.0 & 43.0 & 37.8 & -- & -- & -- & -- \\
SpatialNav~\citep{Zhang2026SpatialNav} & 4.2 & 73.0 & 68.0 & 53.4 & 7.3 & -- & 39.0 & 28.4 \\
\midrule
\rowcolor{spacevlnPurpleRow}
\multicolumn{9}{@{}l}{\textbf{\textit{Zero-shot}}} \\
NavGPT-CE (GPT-4)~\citep{Zhou2024NavGPT} & 8.4 & 26.9 & 16.3 & 10.2 & -- & -- & -- & -- \\
CA-Nav~\citep{Chen2025CANav} & 7.6 & 48.0 & 25.3 & 10.8 & 10.4 & -- & 19.0 & 6.0 \\
Open-Nav (Gemini-2.5-Pro)~\citep{Qiao2025OpenNav} & 7.3 & 30.0 & 23.0 & 19.9 & -- & -- & -- & -- \\
GC-VLN~\citep{Yin2025GCVLN} & 7.3 & 41.8 & 33.6 & 16.3 & 8.8 & 44.4 & 33.8 & 13.8 \\
GTA(GPT-5.1)~\citep{Li2026GTA} & \second{5.0} & 56.2 & 48.8 & \best{41.8} & 6.3 & -- & 46.2 & \best{39.3} \\
\midrule
\rowcolor{spacevlnOwnRow}
\textbf{SpaceVLN (Qwen3.5 Open-Source)} & \second{5.0} & \second{62.4} & \second{48.9} & 32.0 & \second{6.1} & \second{52.4} & \second{46.6} & 30.1 \\
\rowcolor{spacevlnOwnRow}
\textbf{SpaceVLN (Qwen3.5-Plus/Flash)} & \best{4.8} & \best{66.8} & \best{53.3} & \second{34.5} & \best{6.0} & \best{56.6} & \best{48.9} & \second{31.7} \\
\bottomrule
\end{tabular}
\end{table}

\begin{table}[b!]
\centering
\scriptsize
\setlength{\tabcolsep}{1.25pt}
\renewcommand{\arraystretch}{0.94}
\begin{minipage}[t]{0.49\textwidth}
\centering
\caption{Comparison on GN-Bench Benchmark. TF: training-free; ZS: zero-shot.}
\label{tab:navgbench}
\vspace{-3pt}
\begin{tabular}{@{}L{0.34\linewidth}|C{0.07\linewidth}C{0.07\linewidth}|C{0.10\linewidth}C{0.10\linewidth}C{0.10\linewidth}C{0.10\linewidth}@{}}
\toprule
\textbf{Method} & \textbf{TF} & \textbf{ZS} &
\textbf{NE}$\downarrow$ & \textbf{OSR}$\uparrow$ &
\textbf{SR}$\uparrow$ & \textbf{SPL}$\uparrow$ \\
\midrule
CMA~\citep{Krantz2020VLNCE} & \xmark & \cmark & 8.3 & 15.7 & 12.5 & 11.9 \\
NaVid~\citep{Zhang2024NaVid} & \xmark & \xmark & 7.4 & 19.4 & 18.8 & 18.8 \\
Uni-NaVid (ZS)~\citep{Zhang2025UniNaVid} & \xmark & \cmark & 7.9 & 22.2 & 15.0 & 12.5 \\
Uni-NaVid~\citep{Zhang2025UniNaVid} & \xmark & \xmark & 7.2 & 24.1 & 22.5 & 21.9 \\
InternNav (ZS)~\citep{Wei2026DualVLN} & \xmark & \cmark & 7.4 & 23.1 & 18.8 & 17.5 \\
InternNav~\citep{Wei2026DualVLN} & \xmark & \xmark & 7.1 & 22.5 & 22.4 & 22.4 \\
GN-BAE~\citep{gn0_2026} & \xmark & \xmark & 4.9 & 49.5 & 46.9 & 45.1 \\
\midrule
\rowcolor{spacevlnBlueRow}
\textbf{SpaceVLN} & \cmark & \cmark & \textbf{5.8} & \textbf{51.2} & \textbf{39.3} & \textbf{24.2} \\
\bottomrule
\end{tabular}
\end{minipage}\hfill
\begin{minipage}[t]{0.49\textwidth}
\centering
\caption{Comparison on HM3D-OVON Val-Unseen. \textbf{Best} and
\underline{second} results are highlighted.}
\label{tab:ovon}
\vspace{-3pt}
\begin{tabular}{@{}L{0.52\linewidth}|C{0.11\linewidth}|C{0.13\linewidth}C{0.13\linewidth}@{}}
\toprule
\textbf{Method} & \textbf{TF} &
\textbf{SR}$\uparrow$ & \textbf{SPL}$\uparrow$ \\
\midrule
Uni-NaVid~\citep{Zhang2025UniNaVid} & \xmark & 39.5 & 19.8 \\
NavFoM~\citep{Zhang2026NavFoM} & \xmark & 45.2 & 31.9 \\
VLFM~\citep{Yokoyama2024VLFM} & \cmark & 35.2 & 19.6 \\
TANGO~\citep{Ziliotto2025TANGO} & \cmark & 35.5 & 19.5 \\
MetaNav~\citep{Li2026MetaNav} & \cmark & 46.1 & 29.8 \\
MSGNav~\citep{Huang2026MSGNav} & \cmark & 48.3 & 27.0 \\
DRIVE-Nav~\citep{Gao2026DRIVENav} & \cmark & \second{50.2} & \second{32.6} \\
\midrule
\rowcolor{spacevlnBlueRow}
\textbf{SpaceVLN} & \cmark & \best{51.6} & \best{33.1} \\
\bottomrule
\end{tabular}
\end{minipage}
\end{table}

Table~\ref{tab:r2rce_rxrce_merged} compares SpaceVLN with state-of-the-art
VLN-CE methods under supervised learning, pre-exploration, and zero-shot
settings. 
SpaceVLN establishes a new zero-shot SOTA with 53.3 SR on R2R-CE and 48.9 SR
on RxR-CE, surpassing the previous SOTA GTA with relative SR gains of 9.2\%
and 5.8\%, respectively.
Notably, despite requiring no task-specific training or pre-exploration,
SpaceVLN outperforms several supervised or pre-exploration baselines.
On GN-Bench~\citep{gn0_2026}, Table~\ref{tab:navgbench} shows that our
training-free method outperforms many training-based baselines and achieves the
best zero-shot 39.3 SR.

\begin{table}[t!]
\centering
\scriptsize
\setlength{\tabcolsep}{2.0pt}
\renewcommand{\arraystretch}{0.92}
\caption{Key components ablation of SpaceVLN on R2R-CE. HSM: Hierarchical Spatial Memory; L-LM: Local Landmark Memory; P/E-SCoT: planner/executor Spatial-CoT; SC: Stagewise Control.}
\label{tab:ablation}
\begin{tabular}{@{}L{0.285\textwidth}|*{2}{C{0.052\textwidth}}|*{3}{C{0.066\textwidth}}|C{0.047\textwidth}C{0.054\textwidth}C{0.054\textwidth}C{0.054\textwidth}C{0.060\textwidth}@{}}
\toprule
\multirow{2}{=}{\textbf{Method}} &
\multicolumn{2}{c|}{\begin{tabular}{@{}c@{}}\textcolor{spacevlnGreenText}{\textbf{Spatial Memory}}\\[-3.0pt]\rule{0.104\textwidth}{0.35pt}\end{tabular}} &
\multicolumn{3}{c|}{\begin{tabular}{@{}c@{}}\textcolor{spacevlnBlueText}{\textbf{Spatial Reasoning / Stagewise}}\\[-3.0pt]\rule{0.196\textwidth}{0.35pt}\end{tabular}} &
\multicolumn{5}{c@{}}{\begin{tabular}{@{}c@{}}\textbf{R2R-CE Val-Unseen}\\[-3.0pt]\rule{0.270\textwidth}{0.35pt}\end{tabular}} \\
&
\textbf{HSM} & \textbf{L-LM} &
\textbf{P-SCoT} & \textbf{E-SCoT} & \textbf{SC} &
\textbf{NE}$\downarrow$ & \textbf{OSR}$\uparrow$ & \textbf{SR}$\uparrow$ & \textbf{SPL}$\uparrow$ & \textbf{Steps}$\downarrow$ \\
\midrule
\rowcolor{spacevlnGreenRow}
\multicolumn{11}{@{}l}{\textbf{\textit{Ablation of Hierarchical Spatial Memory and Local Landmark Memory}}} \\
w/o HSM+L-LM & \xmark & \xmark & \cmark & \cmark & \cmark & 6.6 & 57.4 & 37.3 & 23.6 & 185.7 \\
w/o HSM & \xmark & \cmark & \cmark & \cmark & \cmark & 6.0 & 58.6 & 38.9 & 24.9 & 185.0 \\
w/o L-LM & \cmark & \xmark & \cmark & \cmark & \cmark & 5.6 & 59.0 & 42.9 & 27.1 & 184.1 \\
\midrule
\rowcolor{spacevlnBlueRow}
\multicolumn{11}{@{}l}{\textbf{\textit{Ablation of Task-Guided Spatial Reasoning and Stagewise Control}}} \\
w/o P-SCoT+E-SCoT+SC & \cmark & \cmark & \xmark & \xmark & \xmark & 5.4 & 58.1 & 38.3 & 21.9 & 205.5 \\
w/o P-SCoT+SC & \cmark & \cmark & \xmark & \cmark & \xmark & 5.6 & 58.0 & 39.4 & 23.5 & 205.3 \\
w/o P-SCoT+E-SCoT & \cmark & \cmark & \xmark & \xmark & \cmark & 5.6 & 58.6 & 42.5 & 26.1 & 182.8 \\
w/o P-SCoT & \cmark & \cmark & \xmark & \cmark & \cmark & 5.3 & 57.2 & 44.8 & 30.7 & 178.3 \\
w/o E-SCoT & \cmark & \cmark & \cmark & \xmark & \cmark & 5.2 & 60.1 & 47.1 & 31.5 & 175.5 \\
\midrule
\rowcolor{spacevlnOwnRow}
\textbf{Full pipeline} & \cmark & \cmark & \cmark & \cmark & \cmark & \textbf{4.8} & \textbf{66.8} & \textbf{53.3} & \textbf{34.5} & \textbf{171.7} \\
\bottomrule
\end{tabular}
\end{table}

\paragraph{Object-Goal Navigation.}
On HM3D-OVON, SpaceVLN achieves a new SOTA with 51.6 SR and 33.1 SPL,
surpassing DRIVE-Nav with relative SR gains of 2.8\%.
This confirms the effectiveness of spatial memory and reasoning across both
instruction-following and Object-Goal Navigation.

\subsection{Ablation Analysis}
\label{sec:experiment-ablation}

\paragraph{Spatial Cognitive Memory.}
Removing both Hierarchical Spatial Memory (HSM) and Local Landmark Memory
(L-LM) reduces SR from 53.3 to 37.3 and increases the average steps from
171.7 to 185.7, indicating more disorientation and corrective motion
during exploration.
HSM is the dominant contributor: removing it drops SR by
14.4 points, showing that the global Spatial Waypoint topology is critical for
progress localization and stage prediction.
L-LM provides local support, and its removal lowers SR to 42.9 by weakening
local landmark grounding and tracking.

\paragraph{Task-Guided Spatial Reasoning and Stagewise Control.}
Removing P-SCoT, E-SCoT, and Stagewise Control (SC) drops SR from 53.3 to 38.3
and SPL from 34.5 to 21.9, while increasing the average steps to 205.5.
This indicates that schema-guided Spatial-CoT clarifies progress and next-stage prediction, 
while SC improves efficiency and self-correction through stage verification and replanning.
The larger SR drop from removing P-SCoT than E-SCoT (8.5 versus 6.2 points)
highlights the planner's role in spatial verification and next-stage anchor
selection.
\FloatBarrier

\section{Conclusion}
\label{sec:conclusion}

In this paper, we introduced SpaceVLN, a zero-shot embodied navigation agent
that integrates online Spatial Cognitive Memory into foundation-model spatial reasoning. 
SpaceVLN decomposes navigation into verifiable space--landmark stages, enabling
stagewise closed-loop planning and execution across Vision-and-Language Navigation and
Object-Goal Navigation without task-specific policy training.
It builds Spatial Cognitive Memory from explored regions,
traversed paths, and landmark observations, and uses this memory in Spatial-CoT
to localize progress, reason over spatial relations, and predict the next stage. 
Extensive experiments in simulation and real-world deployment
demonstrate the performance and generalizability of SpaceVLN, validating the
roles of Spatial Cognitive Memory and Task-Guided Spatial Reasoning for robust zero-shot navigation.

\section{Limitations}
\label{sec:limitations}

SpaceVLN remains sensitive to noisy waypoint labels, open-vocabulary grounding
errors, and dense relative-turn instructions, which can affect progress
localization and stopping. It also lacks full long-horizon geometric planning,
so large detours or occluded object regions may require repeated replanning.
Finally, repeated foundation-model calls limit efficiency; future work will
study more compact memory, lighter control, and stronger self-correction.




\bibliography{example}  

\clearpage
\appendix
\section{Supplementary Overview}
\label{app:supplementary-organization}

This supplementary material is organized as follows:
\begin{itemize}
    \setlength{\itemsep}{2pt}
    \setlength{\parskip}{0pt}
    \item Appendix~\ref{app:method-overview} provides the SpaceVLN agent
    overview and runtime pipeline.
    \item Appendix~\ref{app:stagewise-policy} specifies the stage-level subtask
    interface.
    \item Appendix~\ref{app:memory-details} details Spatial Cognitive Memory
    construction.
    \item Appendix~\ref{app:output-contracts} details schema-guided
    planner and executor Spatial-CoT reasoning.
    \item Appendix~\ref{app:reliability-details} details SpaceVLN agent
    stability controls.
    \item Appendix~\ref{app:simulator-settings} specifies simulation
    experiment settings.
    \item Appendix~\ref{app:real-robot-details} describes real-robot
    experiment settings and evaluation.
    \item Appendix~\ref{app:runtime-cost} reports runtime and model comparisons.
    \item Appendices~\ref{app:simulation-episode-visualization}--\ref{app:failure-analysis}
    provide simulator and real-world qualitative cases, followed by failure
    analysis.
\end{itemize}

\section{Supplementary Method Details}
\label{app:method-details}

\subsection{Overview}
\label{app:method-overview}

SpaceVLN organizes navigation through a stagewise closed-loop framework
grounded in Spatial Cognitive Memory and Task-Guided Spatial Reasoning. The
planner periodically reasons over panoramic observations and persistent spatial
memory to emit a verifiable subtask, while the executor realizes this subtask
through local visual cues, landmark observations, and obstacle evidence. The detailed
planner--executor VLM context architecture used by this framework is shown in
Fig.~\ref{fig:app-reasoning-pipeline}.

Algorithm~\ref{alg:spacevln-pipeline} summarizes the corresponding runtime
pipeline of the SpaceVLN agent.

\begin{algorithm}[H]
\caption{\textbf{Stagewise Closed-Loop Pipeline of SpaceVLN.}}
\label{alg:spacevln-pipeline}
\footnotesize
\begin{tabularx}{\linewidth}{@{}lX@{}}
\textbf{Require:} & navigation task $q$, online RGB-D observation stream, and
agent pose stream \\
\textbf{Ensure:} & stage sequence $\{\sigma_k\}$, final Spatial Cognitive
Memory, and final status \\
\end{tabularx}
\vspace{1pt}
\begin{tabularx}{\linewidth}{@{}rX@{}}
1 & Initialize $\mathcal{M}_0=\{\mathcal{W}_0,\mathcal{L}_0\}$ with
$\mathcal{W}_0=\{\mathcal{G}_0,\mathcal{P}_0\}$;
$\mathcal{F}_{-1}\leftarrow\emptyset$, $k\leftarrow0$.\\
2 & \textbf{while} the task is incomplete and the episode is active
\textbf{do}\\
3 & \quad $\mathcal{O}^{12}_k\leftarrow\operatorname{LookAround}()$,
$\Psi_k\leftarrow
\operatorname{Serialize}_{\mathrm{HSM}}(\mathcal{G}_k,\mathcal{P}_k,\mu_k)$.\\
4 & \quad $\sigma_k\leftarrow
\operatorname{Validate}\bigl(\pi_{\mathrm{P}}(q,\mathcal{O}^{12}_k,
\Psi_k,\mathcal{F}_{k-1})\bigr)$.\\
5 & \quad Parse
$\sigma_k=(v^{j_k}_k,\mathcal{C}_k,v^{j_k+1}_k,\delta_k,u_k,\ell_k,z_k)$.\\
6 & \quad $(\mathcal{G}_{k+1},\mathcal{P}_{k+1})\leftarrow
\operatorname{Update}_{\mathrm{HSM}}(\mathcal{G}_k,\mathcal{P}_k,v^{j_k}_k)$.\\
7 & \quad \textbf{if} $z_k=1$ or the task goal is verified
\textbf{then} execute \textsc{Stop} and \textbf{break}.\\
8 & \quad $\mathcal{L}^{(k)}_{0}\leftarrow\emptyset$,
$\mathcal{B}^{(k)}_{0}\leftarrow\emptyset$, $t\leftarrow0$.\\
9 & \quad \textbf{repeat}\\
10 & \quad\quad Observe FPV RGB-D input, pose, and obstacle state
$(I_t,D_t,x_t,\Omega_t)$.\\
11 & \quad\quad $\mathcal{L}^{(k)}_{t+1}\leftarrow
\operatorname{Update}_{\mathrm{landmark}}(\mathcal{L}^{(k)}_{t},I_t,D_t,x_t,\ell_k)$.\\
12 & \quad\quad $\mathcal{B}^{(k)}_{t+1}\leftarrow
\operatorname{Compress}_{\mathrm{landmark}}(\mathcal{L}^{(k)}_{t+1})$.\\
13 & \quad\quad Execute
$a_t=\pi_{\mathrm{E}}(I_t,\Omega_t,\mathcal{B}^{(k)}_{t+1},\sigma_k)$ and update
the stage record $\tau_k$; $t\leftarrow t+1$.\\
14 & \quad \textbf{until} arrival, stop, blockage, stage-budget exhaustion, or
episode termination.\\
15 & \quad $\mathcal{B}^{(k)}\leftarrow\mathcal{B}^{(k)}_{t}$,
$\mathcal{F}_k\leftarrow
\Phi(\sigma_k,\tau_k,\mathcal{B}^{(k)})$.\\
16 & \quad $k\leftarrow k+1$.\\
17 & \textbf{end while}\\
\end{tabularx}
\end{algorithm}

The pipeline has two coupled time scales. The outer loop, indexed by the
planning stage $k$, constructs the planner input by serializing Hierarchical
Spatial Memory, namely the Spatial Waypoint graph $\mathcal{G}_k$, executed
Spatial Waypoint chain $\mathcal{P}_k$, and rendered map $\mu_k$, into
$\Psi_k$. The planner produces a validated subtask $\sigma_k$, whose current
anchor updates the persistent Hierarchical Spatial Memory before execution.
The inner
loop, indexed by executor step $t$, is local to this active subtask: Local
Landmark Memory is reinitialized for the selected landmark $\ell_k$, updated
from each FPV RGB-D observation and pose, and compressed into the landmark pool
$\mathcal{B}^{(k)}_t$ for action generation. At stage termination, the final
landmark pool and execution record are summarized as feedback
$\mathcal{F}_k$, closing the loop for verification and replanning.

\subsection{Stagewise Closed-Loop Navigation Interface}
\label{app:stagewise-policy}

Building on Sec.~\ref{sec:method-stagewise}, SpaceVLN uses a shared
stage-level subtask interface for Vision-and-Language Navigation and
Object-Goal Navigation. At each planning boundary, the planner receives the
task, 12-view look-around, enhanced egocentric spatial representation, and
previous-stage feedback. It then emits the subtask $\sigma_k$ in
Eq.~\ref{eq:stage-subtask}, whose executor-facing fields specify the next
anchor, selected panoramic direction, executable instruction, and tracked
landmark.

The selected direction is the 12-view heading that best aligns with the next
anchor while remaining locally traversable. If the selected heading is not
front-facing, the controller first rotates the agent toward it so that
execution starts from a task-aligned view. The executor then chooses among
\textsc{Stop}, \textsc{MoveForward}, \textsc{TurnLeft}, and \textsc{TurnRight}
using the current egocentric RGB-D view, obstacle cues, and the
subtask-specific landmarks maintained by Local Landmark Memory.

Control returns to the planner when the next anchor is verified, \textsc{Stop}
is issued, the local route is blocked, or the stage budget is exhausted. The
execution outcome and stage-end landmark pool are summarized as
feedback $\mathcal{F}_k$ for verification and replanning in the next cycle.

\subsection{Spatial Cognitive Memory Construction}
\label{app:memory-details}

Extending Sec.~\ref{sec:method-memory}, Spatial Cognitive Memory is built from
online RGB-D observations through a metric mapping layer and organized into two
named components: Hierarchical Spatial Memory and Local Landmark Memory.
SpaceVLN first projects RGB-D observations into an online metric map that
records explored free space, obstacles, agent poses, and trajectory history. It
then derives a compact memory state
$\mathcal{M}_t=\{\mathcal{W}_t,\mathcal{L}_t\}$, consistent with the main
method. Here $\mathcal{W}_t=\{\mathcal{G}_t,\mathcal{P}_t\}$ denotes
Hierarchical Spatial Memory, storing the Spatial Waypoint graph
$\mathcal{G}_t$ and the executed Spatial Waypoint chain $\mathcal{P}_t$, while
$\mathcal{L}_t$ denotes Local Landmark
Memory for the current subtask. This
representation provides structured spatial context for high-level planning and
grounded landmark candidates for low-level execution, rather than attempting
dense 3D reconstruction.

\paragraph{Tiled metric RGB-D world map.}
SpaceVLN follows standard RGB-D occupancy mapping practice for embodied
navigation~\citep{Elfes1989Occupancy,Chaplot2020ObjectGoal}. Instead of
allocating a fixed global canvas, it maintains a sparse tiled metric map that
stores only visited tiles. Agent-centered crops are composed on demand for
spatial-memory updates, planner input, and visualization. The metric state
records explored area, obstacles, current pose, trajectory traces, and semantic
channels for mapping categories and subtask landmarks. This expandable map has
the same role as the metric maps used in recent continuous zero-shot navigation
systems~\citep{Chen2025CANav}, but is used here to construct the online Spatial
Waypoint graph and maintain Local Landmark Memory.

\paragraph{Spatial Waypoints and region topology.}
At each planner boundary, the structured planner output provides the current
space--landmark anchor. SpaceVLN instantiates a Spatial Waypoint from this
anchor by parsing it into a normalized region type and local landmark set:
\begin{equation}
\begin{aligned}
    (r_k,\Lambda_k)&=\operatorname{ParseRegion}(v^{j_k}_k),\\
    w_k&=(x_k,f_k,r_k,\Lambda_k).
\end{aligned}
\end{equation}
where $x_k$ is the metric pose and $f_k$ is the active floor identifier. The
new Spatial Waypoint instance is appended to the executed Spatial Waypoint
chain before region-level aggregation,
\begin{equation}
    \mathcal{P}_k=\mathcal{P}_{k-1}\oplus w_k,
\end{equation}
so the executed Spatial Waypoint chain remains available even after multiple
Spatial Waypoint instances are assigned to the same region. Spatial Waypoints
are merged when their region labels and metric supports are consistent, while
connector and floor-transition evidence prevents distinct spaces from being
collapsed across doors, hallways, or stairs.

The Spatial Waypoint graph combines two adjacency sources: map-induced
reachability and the adjacency induced by the executed Spatial Waypoint chain:
\begin{equation}
\begin{aligned}
    \mathcal{G}_t&=(\{w_i\}_{i=1}^{N_t},\mathcal{E}_t),\\
    \mathcal{E}_t
    &=\mathcal{E}^{\mathrm{map}}_t
    \cup \mathcal{E}^{\mathrm{chain}}_t,\\
    \mathcal{E}^{\mathrm{map}}_t
    &=\{(w_i,w_j):\operatorname{Reach}(w_i,w_j;m_t)=1\},\\
    \mathcal{E}^{\mathrm{chain}}_t
    &=\operatorname{Adj}(\mathcal{P}_t).
\end{aligned}
\end{equation}
where $\operatorname{Adj}(\mathcal{P}_t)$ denotes consecutive Spatial Waypoint
pairs in the executed Spatial Waypoint chain. The map-induced adjacency records
local traversability inferred from the current metric map,
whereas the chain-induced adjacency preserves the executed route through the
partially mapped environment.

\paragraph{Planner serialization and map rendering.}
The planner receives an enhanced egocentric spatial representation $\Psi_k$
rather than the full metric map. Before each planner call, SpaceVLN serializes
the agent-centered adjacency list with known connectivity paths, executed
Spatial Waypoint chain, and rendered map:
\begin{equation}
\begin{aligned}
    \Psi_k
    &=\operatorname{Serialize}
    \left(\operatorname{Arrange}_{\mathcal{G}_k}(w_k;w_0,w_{k-1}),
    \mathcal{P}_k,\mu_k\right).
\end{aligned}
\end{equation}
$w_k$, $w_0$, and $w_{k-1}$ denote the current, initial, and previous Spatial
Waypoints. $\operatorname{Arrange}_{\mathcal{G}_k}$ merges overlapping
Spatial Waypoints and orders all Spatial Waypoints around $w_k$ into an
agent-centered adjacency list with known connectivity paths; $\mathcal{P}_k$
and $\mu_k$ provide the executed Spatial Waypoint chain and rendered map. This
mirrors the current-centered retrieval used in graph-based language navigation
systems~\citep{Rana2023SayPlan,Werby2024HOVSG,Zhang2026SpatialNav,Zhou2025FSRVLN},
but the graph is constructed online from embodied RGB-D observations rather
than from a pre-built scene graph.

\paragraph{Local Landmark Memory.}
Local Landmark Memory is subtask-centric. When the planner selects the tracked
landmark $\ell_k$, SpaceVLN normalizes it into grounding queries
$\mathcal{Q}^{\ell}_k=\operatorname{Norm}(\ell_k)$ and runs open-vocabulary
grounding/segmentation only for these active terms using GroundingDINO and
RepViT-SAM~\citep{Liu2023GroundingDINO,Wang2023RepViTSAM}.
Detections are projected with depth and fused by category, metric proximity,
and confidence. Small masks, duplicate masks, and stale detections from
previous subtasks are filtered so that the executor receives only landmark
candidates relevant to the active stage. Opening-like targets such as doorways,
entrances, passages, and stairs are marked by category and depth-profile cues,
and arrival is evaluated with a type-specific threshold rather than a single
object-distance rule.

Before each executor query, landmark instances are ranked and compressed as:
\begin{equation}
\begin{split}
    \mathcal{B}^{(k)}_t =
    \operatorname{TopK}_{\gamma\downarrow,d\uparrow}
    \{(c_i,\bar{x}_i,d_i,\beta_i,\gamma_i,
    \chi^{\mathrm{vis}}_{i,t},\chi^{\mathrm{arr}}_{i,t}):
    c_i\in\mathcal{Q}^{\ell}_k\}.
\end{split}
\end{equation}
where $d_i$ and $\beta_i$ are the distance and bearing from the current pose,
and $\gamma_i$ is the fused confidence. The binary visibility flag
$\chi^{\mathrm{vis}}_{i,t}$ indicates whether the instance is visible in the
current view, while $\chi^{\mathrm{arr}}_{i,t}$ records whether it satisfies the
active landmark query under its type-dependent arrival radius. The executor uses
this compact pool together with the egocentric view and obstacle distances for
local action generation. At the next stage boundary, the pool is summarized as
feedback $\mathcal{F}_k$ and reinitialized around the newly selected landmark.

\subsection{Task-Guided Spatial Reasoning Details}
\label{app:output-contracts}

Given the memory state described above, SpaceVLN uses two schema-guided
Spatial-CoT variants for task-guided reasoning, one for stage-level planning
and one for local execution. The planner variant operates at planning
boundaries and produces a structured stage subtask, whereas the executor
variant operates at high frequency to produce primitive actions under the
current subtask. Together, these schemas constrain frozen instruction-tuned
VLMs to expose intermediate spatial reasoning and produce parseable outputs,
without requiring navigation-specific reasoning-trace supervision or free-form
transcripts. This design follows zero-shot and structured CoT
variants~\citep{Wei2022CoT,Kojima2022ZeroShotCoT,Li2023StructuredCoT}, while
differing from methods that learn disentangled navigation reasoning
traces~\citep{Lin2025NavCoT}.

\begin{figure}[t]
    \centering
    \includegraphics[width=\linewidth]{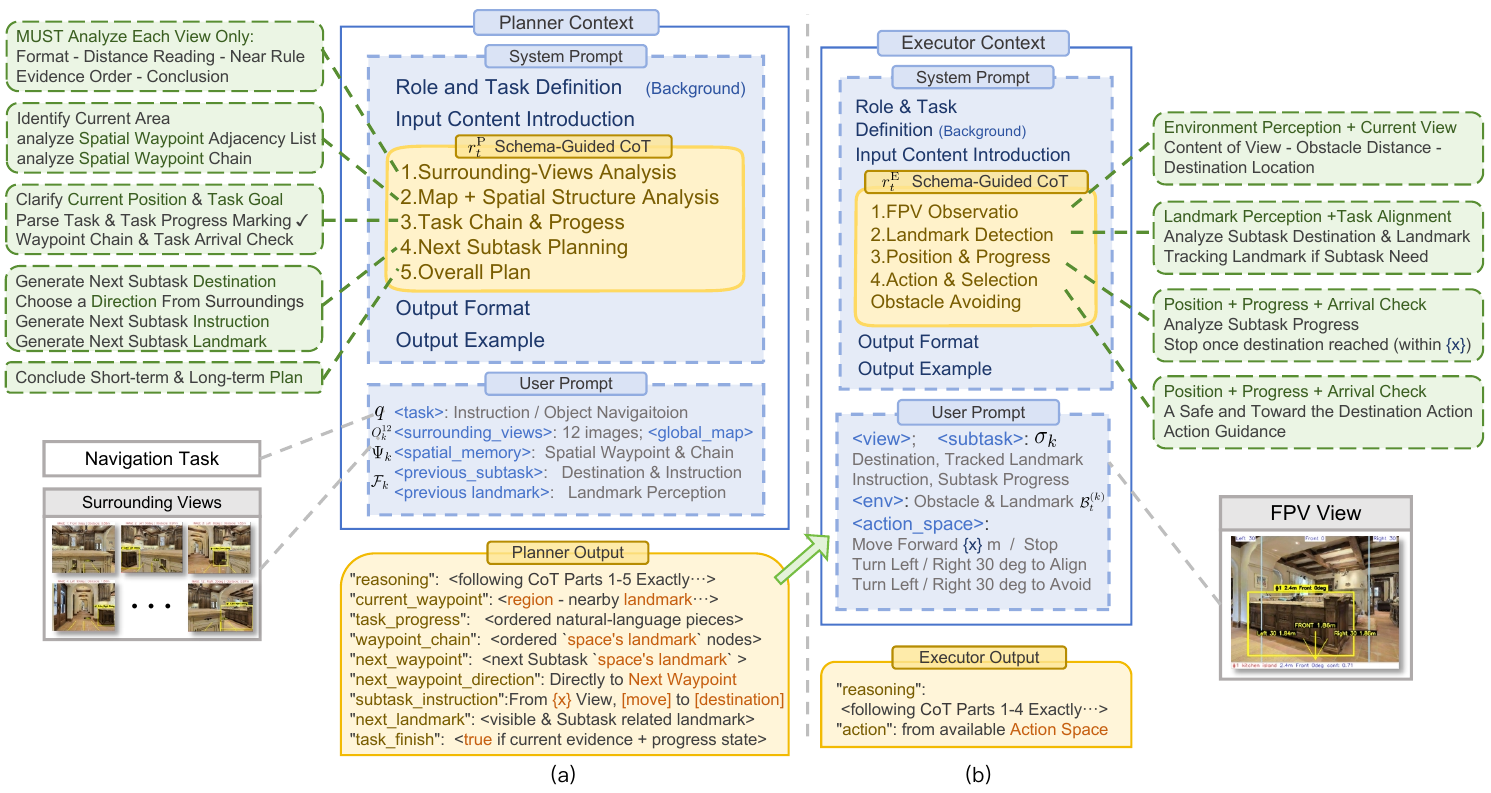}
    \caption{\textbf{VLM Context Architecture of SpaceVLN.}
    (a) The planner context is assembled from the task goal, 12-view
    panoramic observations, the enhanced egocentric spatial representation, and
    previous-stage feedback, providing the input to planner-side Spatial-CoT
    for progress localization and next-stage generation. (b) The executor
    context combines the current subtask, FPV observation, Local Landmark
    Memory, and obstacle cues, providing the input to executor-side Spatial-CoT
    for primitive action generation.}
    \label{fig:app-reasoning-pipeline}
\end{figure}

\paragraph{Planner Spatial-CoT.}
At each planning boundary, the planner receives the task, the 12-view
look-around, the enhanced egocentric spatial representation $\Psi_k$, and the
previous-stage feedback. As shown in
Fig.~\ref{fig:app-reasoning-pipeline}(a), rather than generating a route from
the task alone, the planner uses this context to localize task progress. It
first parses the instruction into an ordered space--landmark anchor chain,
aligns this chain with the current panoramic evidence and $\Psi_k$, and
incorporates previous-stage feedback to determine whether the last subtask has
been completed, should continue, or requires a transition to the next unfinished
stage.

The planner-side Spatial-CoT uses five fields: observation analysis, perception
grounded in Spatial Cognitive Memory, progress estimation, next-stage planning,
and plan summarization. These fields identify visible spaces and landmarks,
relate them to Spatial Cognitive Memory, verify the ordered task chain, and
serialize the next stage-level subtask. The resulting planner output follows
Eq.~\ref{eq:stage-subtask}. It is accepted only when the selected direction
belongs to the provided 12-view look-around and the proposed next anchor is
consistent with the ordered task chain.

\paragraph{Executor Spatial-CoT.}
The executor uses a compact Spatial-CoT for fast local control. As shown in
Fig.~\ref{fig:app-reasoning-pipeline}(b), it follows the current subtask and
local perception to infer a feasible primitive action. Its input context is:
\begin{equation}
    \mathcal{I}^{\mathrm{E}}_t=(I_t,\Omega_t,\Lambda_t,\sigma_k),
\end{equation}
where $I_t$ is the egocentric image, $\Omega_t$ summarizes local obstacle state,
$\Lambda_t$ is Local Landmark Memory for the current subtask, and $\sigma_k$ is
the planner-issued subtask. The executor-side schema contains four fields:
front-view content, landmark validation, stage progress, and obstacle-aware
action generation. The output is parsed into an admissible primitive action:
\begin{equation}
    r_t^{\mathrm{E}}=
    [r_{\mathrm{view}},r_{\mathrm{det}},r_{\mathrm{prog}},r_{\mathrm{act}}],
    \qquad
    a_t=\operatorname{Parse}_{\mathcal{U}_t}(r_t^{\mathrm{E}}),
\end{equation}
where $\mathcal{U}_t\subseteq\mathcal{A}$ is the locally admissible action set.
Arrival checks and action constraints are handled by Local Landmark Memory and
the reliability mechanisms in Appendix~\ref{app:reliability-details}.

\subsection{Control Reliability}
\label{app:reliability-details}

Control reliability keeps the SpaceVLN agent stable under imperfect VLM
service responses and embodied uncertainty. At runtime,
SpaceVLN validates stage-level model outputs, constrains executor actions by
local traversability, monitors realized motion and turn oscillations, and
terminates stages using landmark-grounded arrival evidence together with
stage-level and episode-level budgets. These checks preserve the common stage
interface in Section~\ref{sec:method-stagewise}: invalid model outputs trigger
retry or controlled rejection, unsafe or ineffective actions become recovery or
replanning signals, and budget exhaustion yields a controlled handoff or stop.
Threshold values and benchmark-specific budgets are reported in the
experimental setup.

\paragraph{Model-output validation.}
Before a VLM response is used by the controller, SpaceVLN validates that it can
be parsed into the typed variables required by the stage interface. For the
planner, a response is accepted only if all required fields are present, the
response can be parsed into valid stage variables, and the selected direction
matches one of the provided panoramic image labels. Invalid responses trigger a
bounded retry procedure. If the retry budget $N_{\mathrm{retry}}$ is exhausted,
the episode ends in a controlled state rather than passing an ill-formed output
to the controller. Thus, parsing errors, missing fields, invalid image labels,
and unresolved placeholders become bounded retry or termination events.

\paragraph{Action-space constraints.}
The primitive executor action set $\mathcal{A}$ is defined in the stage
interface above. \textsc{MoveForward}$(\rho)$ requests a bounded forward
displacement, while \textsc{TurnLeft}$(\theta)$ and
\textsc{TurnRight}$(\theta)$ request yaw changes. Turn actions serve two
different purposes: target alignment with the planner-specified direction or
tracked landmark, and obstacle avoidance when the current heading is not
traversable. Each parameterized command is decomposed by the low-level
controller into simulator primitive steps or platform-specific motion targets;
concrete control realizations are reported in
Appendices~\ref{app:simulator-settings} and~\ref{app:real-robot-details}.

The admissible action set $\mathcal{U}_t\subseteq\mathcal{A}$ is updated from
realized motion rather than from the VLM output alone. Ineffective forward
motion temporarily masks \textsc{MoveForward} from the next executor query,
while excessive consecutive turns are limited to avoid oscillatory local
control; we use $N_{\mathrm{turn}}=3$ in the reported experiments. If the
executor generates an action outside $\mathcal{U}_t$, it is re-queried with the
constrained action set. If no admissible progress action remains before
arrival, the stage is reported as locally blocked and control returns to the
planner.

\paragraph{Stage and Episode Termination Criteria.}
Executor rollout for a subtask stops when the executor explicitly generates
\textsc{Stop} or when the subtask destination is reached according to the
Local Landmark Memory arrival flag. The arrival check is tied to the tracked
landmark of the current next anchor, so reaching an unrelated visible landmark
does not terminate the stage. When stage arrival is detected, the controller
returns control to the planner for next-stage verification.

Episode-level stopping is stricter. A final \textsc{Stop} is accepted only when
the final anchor is localized, all earlier anchors in the task chain have been
verified, and the final anchor is grounded by landmark-arrival evidence. If the
executor does not explicitly stop after reaching the goal, stable final arrival
can also terminate the episode when the agent remains inside the final region
with negligible motion. Stage and episode budgets are tracked separately: stage
budget exhaustion returns control to the planner with the current motion and
landmark evidence, while insufficient remaining episode budget yields a
controlled termination and metric recording.

\FloatBarrier

\section{Supplementary Experimental Details}
\label{app:experiment-setup}

\subsection{Simulation Experiments}
\label{app:simulator-settings}
\label{app:metrics-reporting}
\label{app:additional-benchmark-results}

\paragraph{Benchmarks.}
We evaluate SpaceVLN on R2R-CE Val-Unseen (1,839 episodes), the en-US subset of
RxR-CE Val-Unseen (2,433 episodes), the complex-instruction subset of GN-Bench
(1,000 episodes), and the official HM3D-OVON Val-Unseen split (3,000 episodes).
For VLN-style continuous navigation, we follow recent zero-shot VLN evaluation
settings~\citep{Qiao2025OpenNav,Chen2025CANav}. For OVON, we follow the official
HM3D-OVON benchmark and the evaluation setting adopted by
DRIVE-Nav~\citep{Yokoyama2024HM3DOVON,Gao2026DRIVENav}.

\paragraph{Metrics.}
For R2R-CE, RxR-CE, and GN-Bench, we report NE, OSR, SR, and SPL following
standard VLN-CE evaluation~\citep{Anderson2018VLN,Krantz2020VLNCE,Ku2020RxR,gn0_2026}.
For the VLN-style benchmarks, $r=3$\,m. For HM3D-OVON, we report SR and SPL from
the standard ObjectNav evaluator, where success requires issuing \textsc{Stop}
within 1\,m of a goal object~\citep{Yokoyama2024HM3DOVON,Gao2026DRIVENav}. NE
is shown in meters, OSR/SR/SPL are shown as percentages, and all values are
rounded to one decimal place. The ablation table additionally reports average
primitive steps as an efficiency diagnostic. Baselines are retained only under
comparable settings: Table~\ref{tab:r2rce_rxrce_merged} separates supervised
learning, pre-exploration, and zero-shot methods, and task-specific observation
or diagnostic columns are omitted when they are not part of the common
comparison.

\paragraph{Deployment.}
All simulator benchmarks use the same SpaceVLN stagewise closed-loop
framework, online Spatial Cognitive Memory construction, and Task-Guided
Spatial Reasoning. The simulator interface provides RGB-D observations
($640\times480$, $79^\circ$ HFOV) and primitive actions
\{\textsc{Stop}, \textsc{MoveForward}, \textsc{TurnLeft}, \textsc{TurnRight}\}.
Each forward action moves 0.25\,m, and each turn rotates $30^\circ$. The episode
budget is 260 primitive steps for VLN-CE, 300 for the complex-instruction
GN-Bench setting, and 280 for Object-Goal Navigation. At each planning boundary,
the planner collects a 12-view panoramic look-around by sampling headings every
$30^\circ$; the executor then acts from the egocentric RGB-D stream under the
selected stage subtask.

\subsection{Real-Robot Experiments}
\label{app:real-robot-details}

\begin{wrapfigure}{r}{0.48\linewidth}
    \vspace{-8pt}
    \centering
    \includegraphics[width=\linewidth,height=0.45\textheight,keepaspectratio]{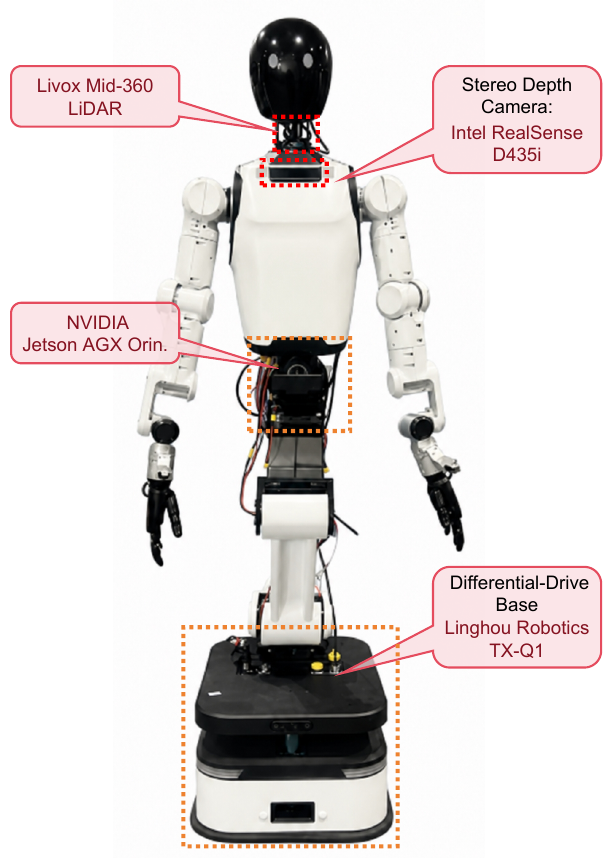}
    \caption{\textbf{Real-robot platform.}
    TX-Q1 mobile base with RealSense D435i RGB-D sensing, Livox Mid-360 LiDAR
    pose input, and Jetson AGX Orin onboard computation.}
    \label{fig:app-real-robot-platform}
    \vspace{-24pt}
\end{wrapfigure}

\paragraph{Hardware Details.}
The real-robot platform is a TX-Q1 differential-drive mobile base developed by
Linghou Robotics. It is equipped with an Intel RealSense D435i RGB-D camera for
visual observation, a Livox Mid-360 LiDAR for pose estimation, and an NVIDIA
Jetson AGX Orin for onboard computation. This configuration provides
synchronized visual, depth, and pose streams for online spatial memory
construction in physical environments.

\paragraph{Real-Robot Deployment.}
On the physical robot, SpaceVLN retains the same high-level agent architecture
as in simulation while adapting sensing and actuation to the platform. The
simulated 12-view panorama is replaced by an 8-view scan collected by stopping
and rotating in 45-degree increments, and synchronized RGB-D observations and
pose estimates are passed to the same spatial memory and reasoning modules. 
For control, the executor uses an adapted action space: target forward
displacements range from 0.5 to 1.5\,m, and yaw rotations are issued in fixed
$45^\circ$ increments. A ROS2 controller tracks each command with live pose
feedback before returning the execution status to the navigation agent.

\paragraph{Experiment Setup.}
Following recent real-world VLN evaluations~\citep{Qiao2025OpenNav,Dai2026EvoNav},
we evaluate SpaceVLN in two physical areas and three settings: Hall, Office,
and Office--Hall. Hall and Office evaluate single-area instruction following,
whereas Office--Hall evaluates cross-area navigation between the two spaces.
The Office area contains dense desks, chairs, and narrow passages, while the
Hall area is relatively open. We define six routes in Hall and six in Office,
each paired with three language variants, yielding 18 instructions per
single-area setting. For Office--Hall, we define five cross-area routes: four
routes have three language variants and one route has two variants, yielding 14
instructions. The real-robot test set contains 50 instructions in total. As in
the real-world setting of Open-Nav~\citep{Qiao2025OpenNav}, an episode is
successful when the robot stops within 2 m of the target.

\begin{table}[!htbp]
\centering
\scriptsize
\setlength{\tabcolsep}{1.6pt}
\renewcommand{\arraystretch}{1.06}
\caption{Real-robot evaluation on the 50-episode test set. Hall and Office contain 18 single-area episodes each, Office--Hall contains 14 cross-area episodes, and Overall aggregates all 50 episodes. NE is in meters. SR is reported as percentage, with successful / total episodes shown in parentheses.}
\label{tab:real_robot}
\begin{tabular}{@{}L{0.17\textwidth}|C{0.047\textwidth}C{0.087\textwidth}|C{0.047\textwidth}C{0.087\textwidth}|C{0.047\textwidth}C{0.087\textwidth}|C{0.047\textwidth}C{0.087\textwidth}@{}}
\toprule
\multirow{2}{=}{\textbf{Method}} &
\multicolumn{2}{c|}{\textbf{Hall}} &
\multicolumn{2}{c|}{\textbf{Office}} &
\multicolumn{2}{c|}{\textbf{Office--Hall}} &
\multicolumn{2}{c@{}}{\textbf{Overall}} \\
\cmidrule(lr){2-3}\cmidrule(lr){4-5}\cmidrule(lr){6-7}\cmidrule(lr){8-9}
& \textbf{NE}$\downarrow$ & \textbf{SR}$\uparrow$
& \textbf{NE}$\downarrow$ & \textbf{SR}$\uparrow$
& \textbf{NE}$\downarrow$ & \textbf{SR}$\uparrow$
& \textbf{NE}$\downarrow$ & \textbf{SR}$\uparrow$ \\
\midrule
CMA~\citep{Krantz2020VLNCE} & 3.02 & 28 (5/18) & 3.04 & 28 (5/18) & 3.53 & 14 (2/14) & 3.17 & 24 (12/50) \\
BEVBert~\citep{An2023BEVBert} & 2.95 & 33 (6/18) & 3.23 & 22 (4/18) & 3.71 & 14 (2/14) & 3.26 & 24 (12/50) \\
NaVid~\citep{Zhang2024NaVid} & 2.75 & 39 (7/18) & 2.98 & 28 (5/18) & 3.27 & 21 (3/14) & 2.98 & 30 (15/50) \\
Open-Nav~\citep{Qiao2025OpenNav} & 2.64 & 39 (7/18) & 2.83 & 33 (6/18) & 3.09 & 29 (4/14) & 2.83 & 34 (17/50) \\
\midrule
\rowcolor{spacevlnOwnRow}
\textbf{SpaceVLN (ours)} & \textbf{2.23} & \textbf{56 (10/18)} & \textbf{2.36} & \textbf{44 (8/18)} & \textbf{2.68} & \textbf{43 (6/14)} & \textbf{2.40} & \textbf{48 (24/50)} \\
\bottomrule
\end{tabular}
\end{table}

\paragraph{Results Analysis.}
Across the real-robot evaluation in Table~\ref{tab:real_robot}, SpaceVLN
achieves the best overall performance, with the largest gains in the
cross-area Office--Hall setting. This pattern suggests that, when facing
complex multi-space navigation tasks, Spatial Cognitive Memory and
Task-Guided Spatial Reasoning improve navigation by maintaining task progress
as a spatial state and grounding subsequent decisions in accumulated spatial
evidence.

\subsection{Runtime Analysis and Model Comparison}
\label{app:runtime-cost}

\begin{table}[!htbp]
\centering
\scriptsize
\setlength{\tabcolsep}{2.4pt}
\renewcommand{\arraystretch}{1.04}
\caption{Method-level runtime and GPU-memory comparison on R2R-CE using RTX 3090-class hardware. Open-Nav and EvoNav use the Gemini-2.5-Pro setting reported by EvoNav; for SpaceVLN, parentheses indicate the planner--executor model configuration.}
\label{tab:runtime_memory}
\begin{tabular}{@{}L{0.46\textwidth}|C{0.065\textwidth}C{0.065\textwidth}|C{0.105\textwidth}C{0.100\textwidth}@{}}
\toprule
\textbf{Method} & \textbf{OSR}$\uparrow$ & \textbf{SR}$\uparrow$ &
\textbf{Time (min)}$\downarrow$ & \textbf{GPU (GB)}$\downarrow$ \\
\midrule
Open-Nav (Gemini-2.5-Pro) & 30.0 & 23.0 & 14.7 & 12.5 \\
EvoNav (Gemini-2.5-Pro) & 51.0 & 43.0 & 7.6 & 2.4 \\
\midrule
\rowcolor{spacevlnOwnRow}
\textbf{SpaceVLN (Qwen3.5-122B-A10B + Qwen3.5-35B-A3B)} & \second{62.4} & \second{48.9} & \best{3.4} & \best{2.3} \\
\rowcolor{spacevlnOwnRow}
\textbf{SpaceVLN (Qwen3.5-Plus + Qwen3.5-Flash)} & \best{66.8} & \best{53.3} & \second{4.3} & \best{2.3} \\
\bottomrule
\end{tabular}
\end{table}

\paragraph{Runtime Analysis.}
Following the runtime evaluation setting of EvoNav~\citep{Dai2026EvoNav}, we
compare method-level efficiency on R2R-CE under RTX 3090-class hardware. 
Table~\ref{tab:runtime_memory} shows that the accuracy gain of SpaceVLN is not
obtained by a slower online loop; the stagewise closed-loop framework improves
both navigation accuracy and online efficiency.

\begin{table}[!htbp]
\centering
\scriptsize
\setlength{\tabcolsep}{0.6pt}
\renewcommand{\arraystretch}{1.04}
\caption{Foundation-model deployment comparison for SpaceVLN on R2R-CE. Cost denotes cache-aware per-episode inference cost. 
$\dagger$ denotes hosted API runs via Alibaba Cloud DashScope with explicit prompt-cache accounting, 
$\ddagger$ denotes local open-weight deployment, and 
$\S$ denotes the MiMo run whose cost is converted from Xiaomi MiMo API Open Platform pricing~\citep{XiaomiMiMoPlatform2026}.}
\label{tab:deployment_breakdown}
\begin{tabular}{@{}L{0.35\textwidth}|C{0.050\textwidth}C{0.050\textwidth}|C{0.100\textwidth}C{0.112\textwidth}C{0.120\textwidth}C{0.050\textwidth}C{0.105\textwidth}@{}}
\toprule
\textbf{Method} & \textbf{OSR}$\uparrow$ & \textbf{SR}$\uparrow$ &
\textbf{Planner (s)}$\downarrow$ & \textbf{Executor (s)}$\downarrow$ &
\textbf{Episode (min)}$\downarrow$ & \textbf{Step}$\downarrow$ & \textbf{Cost (USD)}$\downarrow$ \\
\midrule
Kimi K2.5$^\dagger$ & 61.4 & 50.4 & 47.29 & 6.43 & 12.16 & 177.3& 0.092\\
MiMo-V2.5 + MiMo-V2-Omni$^\S$ & 52.9 & 41.6 & 12.14 & 3.05 & 4.75  & 183.7 & 0.068\\
Qwen3-VL-Plus + Qwen3-VL-Flash$^\dagger$ & 52.2 & 40.3 & 37.19 & 2.68 & 10.77 & 212.4 & 0.028\\
Qwen3.5-122B-A10B + Qwen3.5-35B-A3B$^\ddagger$ & 62.4 & 48.9 & \best{11.10} & \best{1.94} & \best{3.41} & 173.2 & 0.022 \\
Qwen3.5-Plus + Qwen3.5-Flash$^\dagger$ & \best{66.8} & \best{53.3} & 18.84 & 2.23 & 4.34 & \best{171.7} & \best{0.010} \\
\bottomrule
\end{tabular}
\end{table}

\paragraph{Model Comparison.}
Table~\ref{tab:deployment_breakdown} further compares SpaceVLN with different
foundation-model backends, including Kimi K2.5~\citep{KimiTeam2026KimiK25},
MiMo-V2.5/MiMo-V2-Omni~\citep{XiaomiMiMo2026}, Qwen3-VL~\citep{Bai2025Qwen3VL},
and Qwen3.5~\citep{qwen35blog}. Kimi K2.5 reaches a competitive 50.4 SR, but
incurs the longest runtime at 12.16 minutes per episode, highlighting the
latency cost of using a single strong hosted model without flash model. Qwen3.5-Plus/Flash
achieves the best SR (53.3) and the fewest steps. This result aligns
with the Qwen3.5 report's emphasis on stronger multimodal spatial ability~\citep{qwen35blog},
suggesting that SpaceVLN benefits from backend models with stronger spatial
understanding and inference in visual-spatial contexts. The local open-weight
Qwen3.5 stack is the fastest(3.41 minutes, 48.9 SR).

\section{Qualitative Visualizations}
\label{app:visualization-cases}

The qualitative visualizations analyze two successful episodes with a common
logging format and then diagnose representative failure cases. Each successful
case pairs planner-level Spatial Waypoint progress with executor-side visual,
landmark, and map evidence, making it possible to inspect how a stage-level
decision is verified before the agent advances to the next subtask. The failure
analysis groups unsuccessful episodes by error source and exposes the remaining
limitations of online spatial grounding, control, and progress verification.
These figures are diagnostic visualizations; the quantitative conclusions
remain tied to the benchmark tables in the main paper.

\begin{figure}[t]
    \centering
    \setlength{\abovecaptionskip}{3pt}
    \setlength{\belowcaptionskip}{2pt}
    \includegraphics[width=\linewidth,height=0.92\textheight,keepaspectratio]{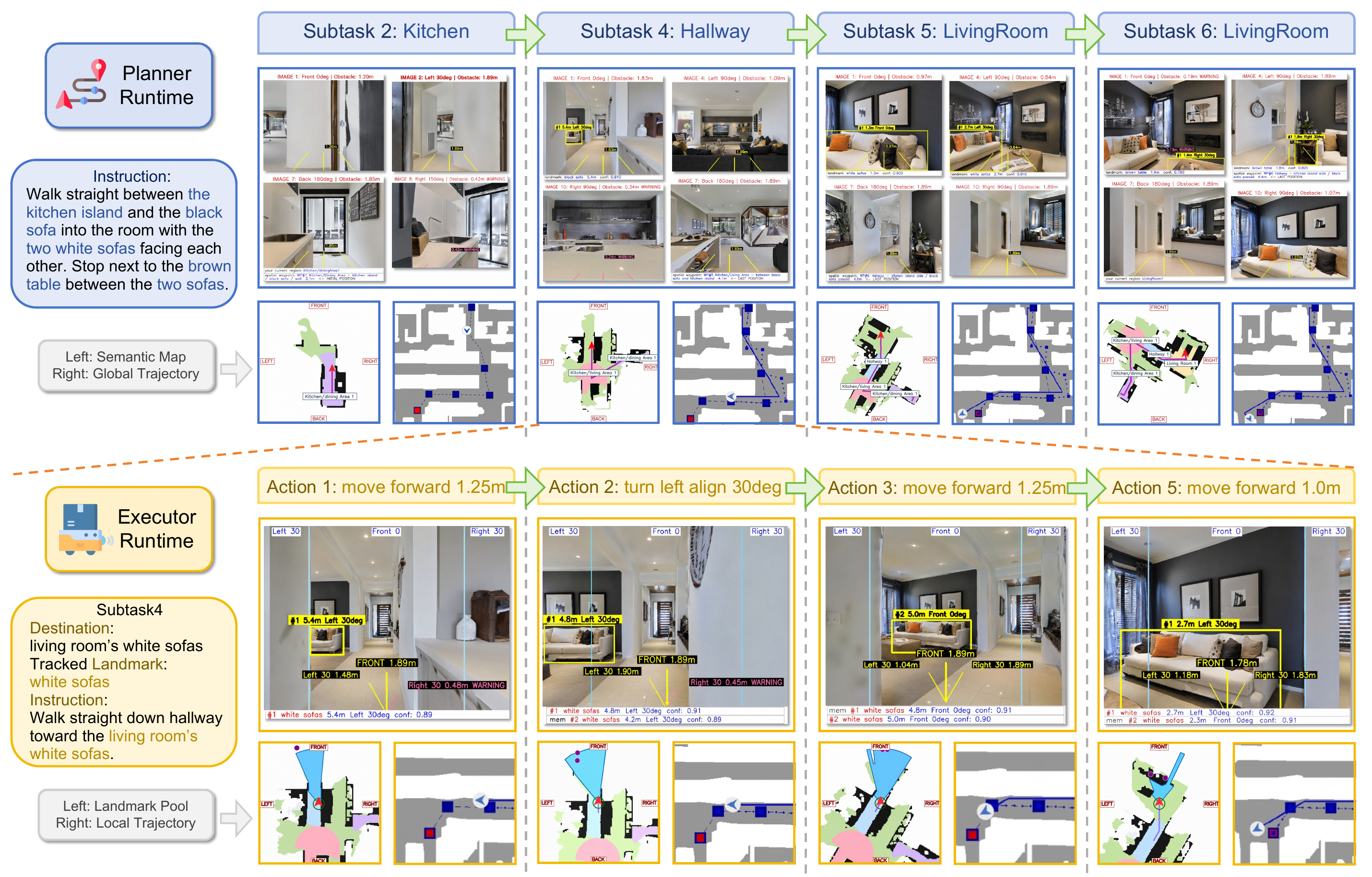}
    \caption{\textbf{Successful simulator episode.}
    The route starts from a hallway, passes around the kitchen area, and enters
    the living room. The planner row shows surrounding views, semantic maps,
    stagewise closed-loop planning decisions, and the top-down global
    trajectory with evaluation markers. The executor row shows one stage
    subtask with FPV observations, Local Landmark Memory, primitive action
    outputs, and the top-down local trajectory.}
    \label{fig:app-simulation-episode}
\end{figure}

\subsection{Simulation Episode Visualization}
\label{app:simulation-episode-visualization}

Figure~\ref{fig:app-simulation-episode} shows a successful simulator episode
in which the agent follows the instruction to pass between the kitchen island
and black sofa, enter the room with two facing white sofas, and stop near the
brown table. This episode requires progress over multiple connected spaces,
from the kitchen area to the hallway and then to the living room. The planner
does not rely on a fixed textual split of the instruction; instead, it
uses the current panoramic observation, semantic map, and accumulated global
trajectory to localize task progress within Spatial Cognitive Memory and select
the corresponding Spatial Waypoints. The executor visualization then shows how
the selected hallway-to-living-room stage is organized around the white-sofa
landmark, with obstacle distances, Local Landmark Memory, and the realized
short trajectory constraining local action generation. This case illustrates
the qualitative advantage of SpaceVLN in complex multi-space navigation: task
progress is maintained as a spatial state, and local actions are constrained by
the current space--landmark stage.

\begin{figure}[t]
    \centering
    \setlength{\abovecaptionskip}{3pt}
    \setlength{\belowcaptionskip}{2pt}
    \includegraphics[width=\linewidth,height=0.92\textheight,keepaspectratio]{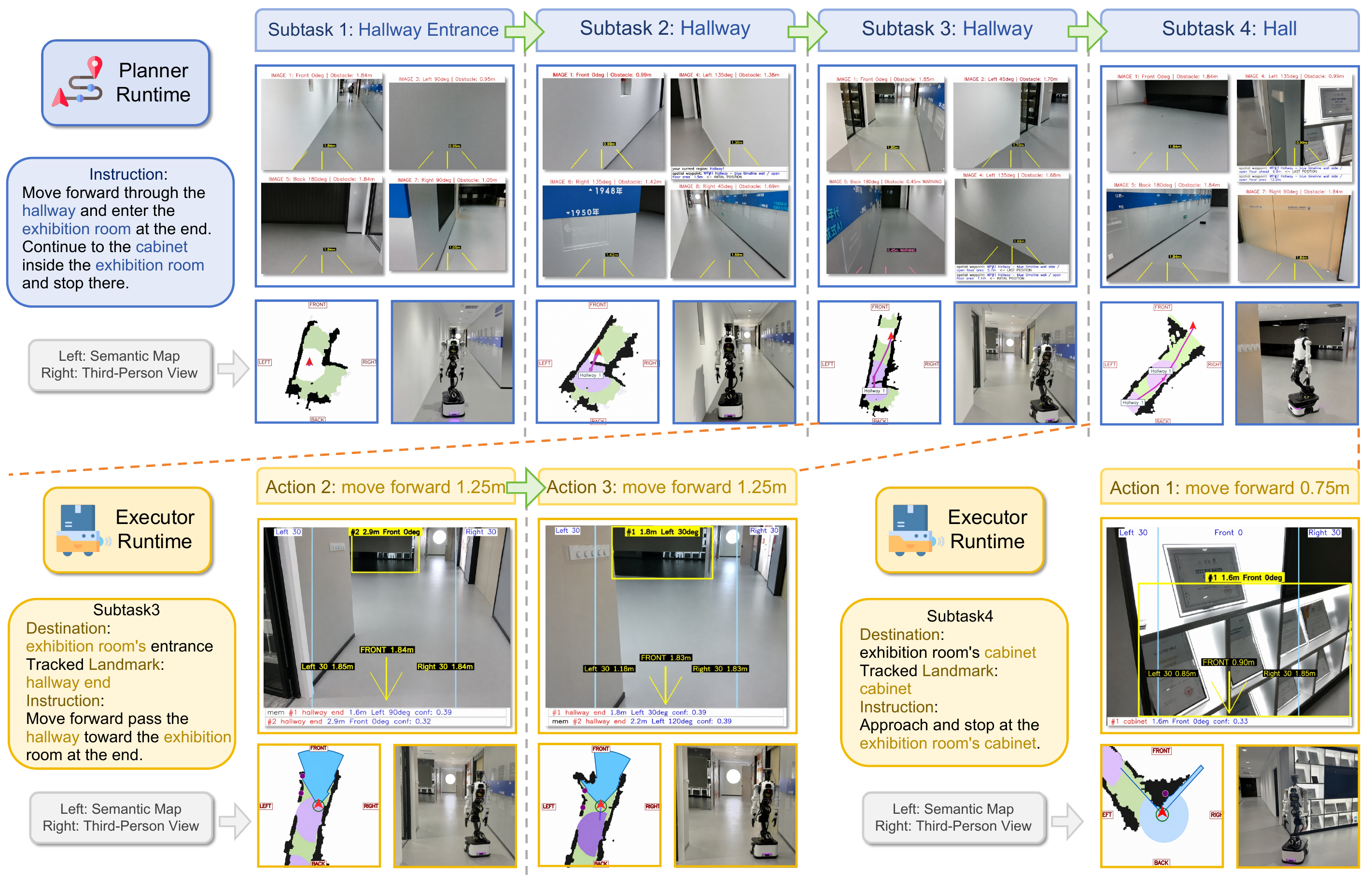}
    \caption{\textbf{Real-robot deployment episode.}
    This real-robot case shows the robot navigating from the hallway into the
    exhibition room and stopping near the cabinet. The planner row contains
    surrounding views, semantic maps, and third-person views; the executor panels
    contain FPV observations, semantic maps, and executed actions for entering the
    exhibition room and approaching the cabinet.}
    \label{fig:app-real-world-episode}
\end{figure}
\subsection{Real-World Episode Visualization}

\label{app:real-world-episode-visualization}

Figure~\ref{fig:app-real-world-episode} applies the same visualization format to
a physical route that starts in a hallway, enters the exhibition room, and
terminates at the cabinet. The planner first advances through 
hallway Spatial Waypoints and then switches to the exhibition-room target,
using selected views and semantic maps; third-person deployment views are
shown only to situate the robot within the physical scene. The executor panels
show two local stages: approaching the exhibition-room entrance from the
hallway end and stopping at the cabinet once it becomes the active landmark.
This case shows that, despite sensing noise, physical constraints, and
platform-induced motion drift, SpaceVLN can maintain spatial progress through
the stagewise closed-loop framework and complete the requested route.

\begin{figure}[t]
    \centering
    \setlength{\abovecaptionskip}{3pt}
    \setlength{\belowcaptionskip}{2pt}
    \includegraphics[width=\linewidth,height=0.82\textheight,keepaspectratio]{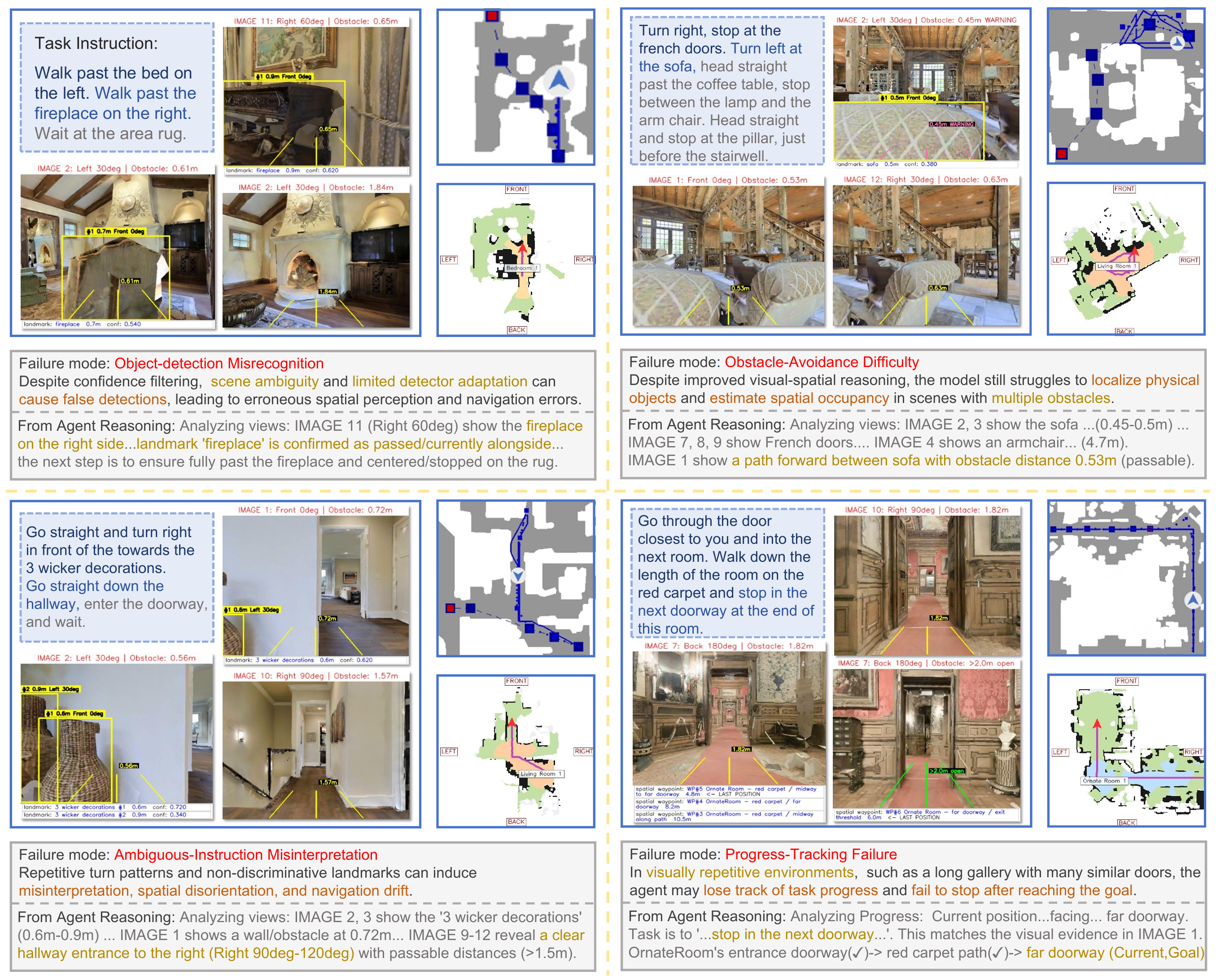}
    \caption{\textbf{Failure cases.}
    Four representative failure modes are shown: object-detection
    misrecognition, obstacle-avoidance difficulty, ambiguous-instruction
    misinterpretation, and progress-tracking failure. Each panel includes the task
    instruction, visual observations, map information, the failure reason, and the
    corresponding agent reasoning excerpt for localizing the source of the error.}
    \label{fig:app-failure-analysis}
\end{figure}

\subsection{Failure Analysis}
\label{app:failure-analysis}

Figure~\ref{fig:app-failure-analysis} summarizes four typical failure modes.
We group these cases by their dominant error source. Object-detection misrecognition
occurs when visually similar or contextually ambiguous landmarks pass
confidence filtering and enter Local Landmark Memory, creating an incorrect
landmark association that can shift the agent's spatial state estimate and
lead the executor to continue toward or stop near the wrong object.
Obstacle-avoidance difficulty
appears in cluttered scenes where the agent must jointly localize the target
and estimate traversable free space;
without a full long-horizon geometric planner, narrow passages, occlusions,
and local detours can still lead to inefficient motion or blocked execution
when the agent cannot find a feasible bypass.
Ambiguous-instruction
misinterpretation arises when relative directions, repeated turns, or weakly
discriminative landmarks destabilize the intended reference frame, causing the
planner to choose an incorrect direction or Spatial Waypoint and leading to
navigation drift. Finally, progress-tracking failure occurs in visually repetitive rooms or corridors,
where multiple doorways or subregions can appear consistent with the same
instruction. In such cases, stage completion may not be recognized in time, so
the agent remains in the current subtask and can continue moving after the
intended goal has already been reached.

These cases are consistent with the limitations discussed in the main paper:
online spatial memory remains sensitive to noisy waypoint labels and
open-vocabulary grounding, dense relative-turn instructions can make progress
localization fragile, and local control lacks complete geometric search over
future detours. Future work should strengthen temporal landmark verification,
maintain explicit reference-frame and progress beliefs, integrate
traversability-aware planning with backtracking, and compress memory so that
these self-correction mechanisms remain efficient in real deployments.

\end{document}